%% file: cas.tex
\documentclass[a4paper,fleqn]{cas-sc}

\usepackage[numbers]{natbib}
\usepackage{color,xcolor}

\definecolor{mygreen}{RGB}{61,145,64}
\usepackage{threeparttable}
\usepackage[edges]{forest}
\usepackage{adjustbox}
\usepackage{tabularx}
\usepackage{multirow}
\usepackage{graphicx}
\usepackage{tabularray}
\usepackage{booktabs}
\usepackage{colortbl}
\usepackage{cleveref}
\usepackage{etoolbox}
\definecolor{ccr}{RGB}{30,144,255} 
\usepackage{hyperref}
\hypersetup{hypertex=true,
    colorlinks=true,
    linkcolor=ccr,
    anchorcolor=ccr,
    citecolor=ccr}

\def\tsc#1{\csdef{#1}{\textsc{\lowercase{#1}}\xspace}}
\tsc{WGM}
\tsc{QE}
\tsc{EP}
\tsc{PMS}
\tsc{BEC}
\tsc{DE}

\begin{document}
\let\WriteBookmarks\relax
\def\floatpagepagefraction{1}
\def\textpagefraction{.001}

\shorttitle{Fake Artificial Intelligence Generated Contents (FAIGC)}

\shortauthors{Xiaomin Yu et~al.}

\title [mode = title]{Fake Artificial Intelligence Generated Contents (FAIGC): A Survey of Theories,
Detection Methods, and Opportunities.}                      

\author[1]{Xiaomin Yu}[orcid=0000-0003-4846-3162]{\fnref{co-first}}
\ead{yuxm02@gmail.com}
\author[1]{Yezhaohui Wang}{\fnref{co-first}}
\ead{yezhaohuiwang@gmail.com}
\affiliation[1]{organization={Cool Large Language Models Research Group},
    city={Beijing}, 
    country={China}}
\fntext[co-first]{Equal Contribution}

\author[2]{Yanfang Chen}{\corref{cor}}
\ead{cyf@ruc.edu.cn}
\author[2]{Zhen Tao}
\ead{taozhen@ruc.edu.cn}
\author[2]{Dinghao Xi}
\ead{xidinghao@ruc.edu.cn}
\author[2]{Shichao Song}
\ead{songshichao@ruc.edu.cn}
\author[2]{Simin Niu}
\ead{niusimin@ruc.edu.cn}
\affiliation[2]{organization={Renmin University of China},
    city={Beijing},
    country={China}}
\cortext[cor]{Corresponding author}

\author[1]{Zhiyu Li}
\ead{aakas888@gmail.com}

\begin{abstract}
In recent years, generative artificial intelligence models, represented by Large Language Models (LLMs) and Diffusion Models (DMs), have revolutionized content production methods. These artificial intelligence-generated content (AIGC) have become deeply embedded in various aspects of daily life and work. However, these technologies have also led to the emergence of Fake Artificial Intelligence Generated Content (FAIGC), posing new challenges in distinguishing genuine information. It is crucial to recognize that AIGC technology is akin to a double-edged sword; its potent generative capabilities, while beneficial, also pose risks for the creation and dissemination of FAIGC. In this survey, We propose a new taxonomy that provides a more comprehensive breakdown of the space of FAIGC methods today. Next, we explore the modalities and generative technologies of FAIGC. We introduce FAIGC detection methods and summarize the related benchmark from various perspectives. Finally, we discuss outstanding challenges and promising areas for future research.
\end{abstract}

% \includegraphics{figs/grabs.pdf}

% Keywords
\begin{keywords}
Fake Information \sep Artificial Intelligence Generated Contents \sep Large Language Model \sep Diffusion Model \sep Deepfake 
\end{keywords}

\maketitle

\input{1_intro}

\input{2_}

\input{3_}

\input{4_}
\input{5_}

\input{6_}

\input{7_}
\input{8_}

\bibliographystyle{cas-model2-names}

\bibliography{cas}

\end{document}

%% file: 1_intro.tex
\section{Introduction}

In recent years, the rapid development and extensive application of generative artificial intelligence models, such as ChatGPT \cite{chatgpt} and Stable Diffusion \cite{ldm}, have gradually transformed our content production methods. AIGC has been deeply integrated into various aspects of our daily lives and professional activities. As depicted in Figure \ref{fig1:examples}(a), conversational AI bots like ChatGPT are now widely employed to assist in content creation, providing creators with significantly helpful outputs characterized by their high authenticity and professionalism. In addition to the automated generation of textual material, as illustrated in Figure \ref{fig1:examples}(b), generative models such as Stable Diffusion have also facilitated the production of images or videos. We can now produce various images or videos that previously did not exist through specific natural language commands or descriptions. The authenticity of these AI-created images and videos has advanced to the point where they are nearly indistinguishable from those created by humans \cite{lu2024seeing,zhu2024genimage,chen2023can}. Moreover, as shown in Figure \ref{fig1:examples}(c), facial forgery technology based on Deepfake \cite{westerlund2019emergence,kwok2021deepfake,chadha2021deepfake} enables the substitution of facial imagery in selected photos or videos with any chosen individual, producing highly realistic outputs. 

\input{Fig_1}

These examples illustrate a critical observation: As AIGC becomes increasingly integrated into our daily lives, education, and work, we must confront the potential risks associated with AIGC, especially the proliferation of false content. For instance, in April 2023, a German magazine published a fabricated interview with Schumacher using AI, which garnered significant online attention. However, Schumacher has not participated in any interviews since his skiing accident in 2013 \cite{Amanda2023}. Additionally, statistics from DeepMedia \cite{Alexandra2023} reveal that the number of online Deepfake videos in 2023 tripled compared to 2022, with an eightfold increase in sophisticated voice forgeries. Recent instances include frauds executed using Deepfake technology \cite{Heather2024}. Some perpetrators trained models using videos of victims' friends posted on social platforms, then crafted fake AI-generated videos by superimposing the friends' faces, thereby creating deceptive video chats to trick the victims. These incidents highlight that the AIGC era facilitates content production and has inadvertently become a powerful tool for creating and spreading misinformation. As depicted in Figure \ref{fig2:plt}, a trend analysis of news keywords related to adverse incidents caused by FAIGC highlights growing public concern. This trend underscores the increased scrutiny and attention directed towards misinformation potentially generated by AIGC, which is parallel with the expanding use and innovation of AIGC.

This trend serves as a reminder that alongside the focus on innovations in AIGC, it is crucial to explore the potential risks associated with FAIGC. Before the widespread advent of AIGC, extensive research was carried out on User-Generated Content (UGC) or Human-Generated Content (HGC). These studies covered a wide range have spanned from sentiment analysis of UGC \cite{yen2019design,serna2016discovery,schmunk2013sentiment,rasool2021reading}, to preference analysis \cite{cheng2010ugc,cheong2008consumers}, and multimodal analysis \cite{shah2016multimodal,shah2017multimodal,pang2015deep}. Researchers from various disciplines have shown profound interest in the natural language, images, videos, and other modalities generated by humans \cite{bahtar2016impact,santos2022so,burgess2009user,daugherty2008exploring,naab2017studies,krumm2008user}. Concurrently, considerable attention was paid to fake content created by humans, including fake news \cite{pennycook2021psychology,jiang2023similarity,lazer2018science,zhang2020overview,gelfert2018fake,kalsnes2018fake,tandoc2019facts}, misinformation \cite{zannettou2019web,giglietto2019fake,kumar2018false}, and rumors \cite{bian2020rumor,chen2020rumor,meel2020fake,choi2020rumor,rani2022rumor}. Processing these falsehoods were developed \cite{bondielli2019survey,ahmad2022efficient,mridha2021comprehensive,roy2018deep}, ranging from text classification models based on statistical machine learning for detecting spam emails \cite{crawford2015survey,makkar2020efficient,kumar2020email,ahmed2022machine} to neural network-based models for rumor identification \cite{ma2018rumor,jin2017multimodal,chen2018call,asghar2021exploring}, and graph neural network-based models for fake news detection \cite{dou2021user,phan2023fake,guo2022mixed,chandra2020graph,mahmud2022comparative}. However, it is essential to recognize that previous research on fake information primarily focused on fake information created manually (Fake-User-GC) \cite{grosser2019trustworthy,saura2020defining,khan2023visual}, with insufficient depth in examining FAIGC. Studies have indicated that the public's ability to detect fake information on social media platforms is relatively limited \cite{roets2017fake}. The emergence of FAIGC in the AIGC era exacerbates this issue. The exceptional capabilities of AIGC models not only significantly reduce the creation and distribution costs of FAIGC but also achieve a high degree of realism, precisely mimicking the style and context of authentic content. This advancement has made disseminating fake information on social media more rapid and covert, challenging the public's discerning and judgment abilities. Thus, developing detection methods for FAIGC is imperative for further study.

\input{Fig_2}

\textbf{Major Contributions.} This survey updated the comprehensive framework for the FAIGC field. In contrast, previous studies have largely focused on verifying whether the content was generated by AI \cite{wu2023survey}, with relatively little exploration into the authenticity of AIGC. Additionally, although there have been initial explorations into the textual modality of FAIGC \cite{chen2023can}, these studies often lacked comprehensive definitions and did not extend to other modalities. In our survey, Chapter 2 addresses three facets of FAIGC: its creative intent, various modalities and generation techniques, and the methods employed in its creation. Chapters 3 and 4 delve into the generation techniques for different and multimodal FAIGC, categorized under AI-generated disinformation and AI-generated misinformation. Chapter 5 presents a detailed classification of FAIGC detection methods across various task types. Chapter 6 provides an overview of benchmarks and datasets relevant to the three selected FAIGC detection tasks. Finally, Chapter 7 offers an in-depth discussion of the prospective research directions and trends in FAIGC. The overarching contributions of this survey are outlined as follows:

\begin{itemize}

  \itemsep 1.2em
  \item \textbf{In-depth and Comprehensive Analysis.} This is the first comprehensive survey that systematically reviews studies on the field of FAIGC. We firmly believe that the taxonomy framework, summaries, and progress we present in this survey can significantly promote research in FAIGC.
  \item \textbf{Innovative and Structured Taxonomy for FAIGC.} Our classification and discussion of FAIGC encompass three distinct dimensions: the intent behind FAIGC, the modalities and generative technologies of FAIGC, and the creation method of FAIGC. This multi-dimensional approach facilitates a thorough and novel understanding of FAIGC.
  \item \textbf{Comprehensive Overview of FAIGC Generative Technologies.} We provide a detailed overview of the principal technologies used to generate various types of FAIGC across text, visual, audio, and multimodal formats, particularly focusing on generative models such as LLMs and DMs.
  \item \textbf{Thorough Survey of FAIGC Detection Methods.} Addressing the unique aspects of FAIGC detection in different modalities, we categorize FAIGC into three sub-tasks: Deception FAIGC Detection, Hallucination-based FAIGC Detection, and Deepfake Detection. The survey elaborates on the existing detection methodologies for these tasks across various modalities.
  \item \textbf{Extensive Compilation of FAIGC-Related Datasets.} Our investigation methodically compiles and classifies existing datasets and benchmarks pertinent to FAIGC. It comprehensively analyzes these resources, focusing on their modalities, types, and suitability for specific tasks.
  \item \textbf{Future Research Directions in FAIGC.} We explore several promising avenues for future research in FAIGC detection. These include diverse areas such as multimodal FAIGC Detection, Zero-shot FAIGC Detection, and interpretable FAIGC Detection and so on. It provides valuable ideas for future researchers in this field.
\end{itemize}

%% file: Fig_1.tex
\begin{figure}[ht]
  \centering
  \includegraphics[width=0.9\textwidth]{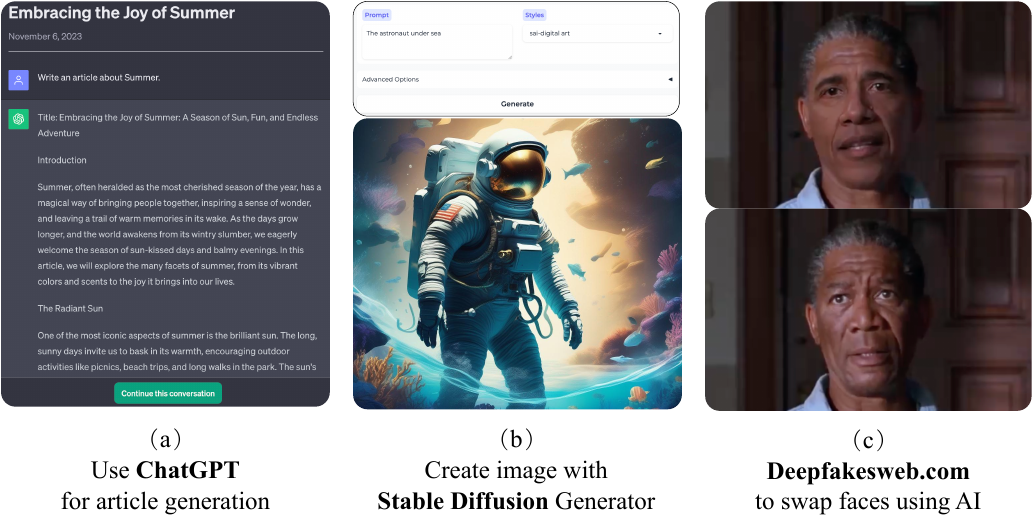} 
  \caption{Typical AIGC examples: text generation, image generation, and deepfake in videos.}
  \label{fig1:examples} 
\end{figure}

%% file: Fig_2.tex
\begin{figure}[ht]
  \centering
  \includegraphics[width=0.92\textwidth]{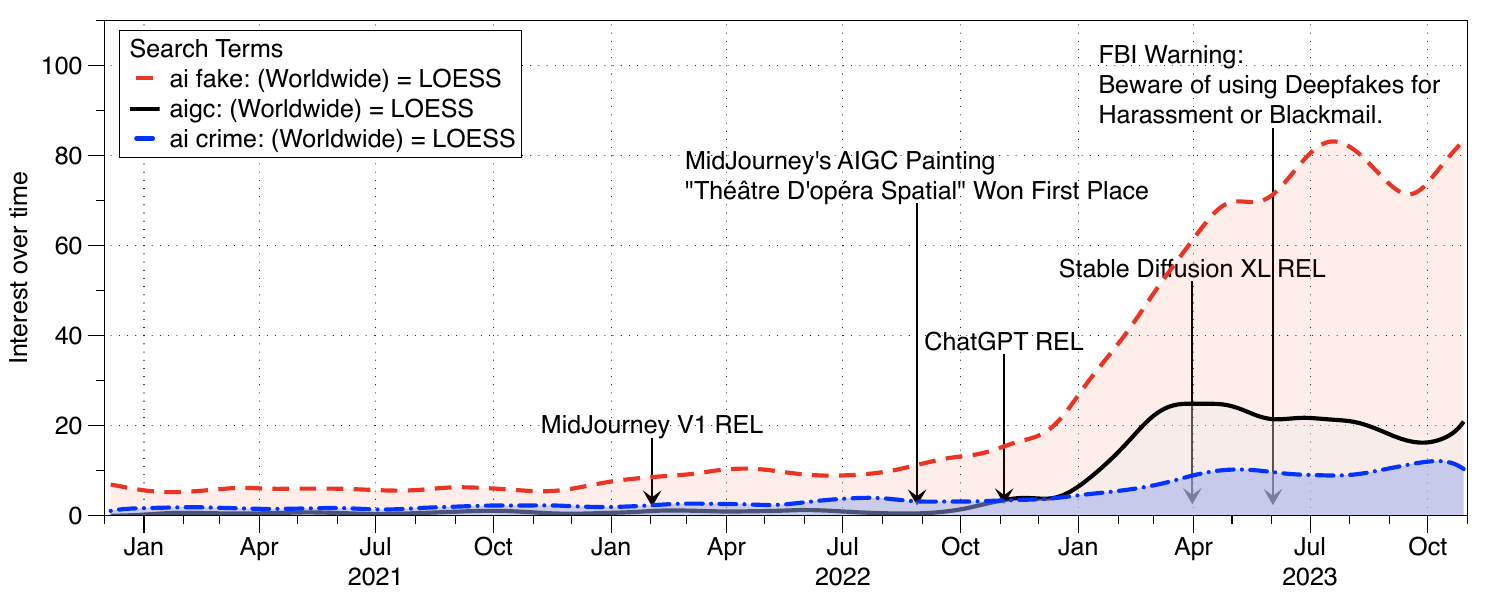} 
  \caption{This line chart clearly illustrates that the search trends related to AI forgery, or AI crime show a strong positive correlation with the search trends for AIGC. This serves as a reminder that while focusing on the innovations brought about by AIGC, we must also further investigate the potential risks associated with FAIGC.}
  \label{fig2:plt} 
\end{figure}

%% file: 2_.tex
\section{FAIGC Taxonomy} 

In this chapter, we examine the current state of research on FAIGC by focusing on three key aspects: the intent behind FAIGC, the modalities and the generative technologies of FAIGC, and the creation method of FAIGC. Our comprehensive taxonomy is depicted in Figure \ref{fig:framework}.

\input{Fig_3}

\subsection{The Intent Behind FAIGC}

Kumar et al. \cite{kumar2018false} taxonomize traditional fake information into two principal types based on the underlying intent: Disinformation and Misinformation. Similarly, we divide FAIGC into two categories based on subjective intention: AI-generated disinformation and AI-generated misinformation.

\textbf{AI-generated Disinformation (AIGD)} refers to fake information created and disseminated with the intent to deceive,  such as hoaxes \cite{kumar2016disinformation}, fake news \cite{lazer2018science,zhou2020survey}, and rumors \cite{friggeri2014rumor}. In the era dominated by AIGC, AIGD resembles traditional false information in form. However, there has been a significant change in the subject of its creation. While traditional fake information is usually concocted and spread by humans, in the AIGC era, the creators are generative models such as LLMs and DMs. AIGD typically originates from users deliberately guiding models to generate fake information through designed prompts. For instance, users might intentionally design prompts based on specific political, economic, or social motives to create biased or entirely fake information. AIGD is characterized by its explicit intent to deceive, aiming to mislead the public, manipulate social viewpoints, or influence public decision-making. Compared to traditional fake information, it potentially has more severe consequences \cite{menczer2023addressing}.

\textbf{AI-generated Misinformation (AIGM)} is attributed to the limitations in the capabilities of generative models, which can result in the creation of inaccurate content that deviates from facts. AIGM arises not from an intent to deceive but from the inherent shortcomings in generative models, which lead to discrepancies between the generated content and factual reality. AIGM are particularly prevalent in texts generated by LLMs and Multimodal Large Language Models (MLLMs), primarily due to hallucination \cite{huang2023survey}. Compared to the harm characterized by AI-generated Disinformation, AIGM primarily manifests as factual inaccuracies. Its potential danger is comparatively lesser due to the lack of intentional malice. However, as the capabilities of generative models continue to improve, the AIGM becomes increasingly convincing, thereby escalating the difficulty in discerning the authenticity of the contents.

\subsection{The Modalities and Generative Technologies of FAIGC}

In this section, we explore the various modalities and generative technologies associated with artificially generated information content (FAIGC) fabrication. We examine the distinct modalities including text, visual, audio, and multimodal forms. Each modality leverages different generative technologies such as LLMs, or DMs.

\textbf{Text Modality.} Text modality is the most common form in FAIGC. In the text modality, FAIGC is primarily generated by LLMs, such as the Llama series \cite{touvron2023llama,touvron2023llama2}, Mistral \cite{jiang2023mistral}, and so on. The malicious people generate FAIGC by constructing prompts in ways like "Controllable Generation" and "Arbitrary Generation" (Sec. \ref{sec3:text modality}). Compared to humans, the high-efficiency generation characteristic of LLMs has significantly increased the quantity and severity of FAIGC \cite{mozes2023use}. Studies \cite{chen2023can, spitale2023ai} have found that FAIGC generated by LLMs (such as ChatGPT) is more difficult to detect than human-written content with similar semantics. This shows that FAIGC generated by LLMs might be presented in more deceptive ways. Moreover, to generate harmful content, some individuals use "Jailbreak" technology \cite{jiang2024artprompt, liu2023autodan, jones2023automatically} to circumvent the security protocols of LLMs. This practice has caused researchers serious concern about security threats and vulnerabilities associated with LLMs \cite{mozes2023use, kang2023exploiting}.

\textbf{Visual Modality.} The FAIGC in visual modality includes both images and videos. Images, typically do not inherently contain a right or wrong nature. Even if an image is generated by AI, it usually cannot be categorized as FAIGC. The image's authenticity often requires additional context or fact-checking for verification. As for videos, the most common approach involves using Deepfake technology \cite{yu2021survey,malik2022deepfake} to create fraudulent \cite{weikmann2023visual} or malicious videos \cite{mania2024legal}. FAIGC in visual modality primarily leverages DMs. The generation within DMs typically involves two processes: a forward process that incrementally degrades data by adding noise, and a reverse process that learns to generate new data by denoising \cite{yang2022diffusion}. Current research on diffusion models primarily focuses on three main methods: Denoising Diffusion Probabilistic Models (DDPMs) \cite{ho2020denoising}, Score-based Generative Models (SGMs) \cite{song2019generative}, and Score-based Stochastic Differential Equations (Score-based SDEs) \cite{song2020score}. Through DMs, the generation \cite{yang2024mastering,harvey2022flexible} and editing \cite{huang2024diffusion} of images and videos are accomplished.

\textbf{Audio Modality.} FAIGC often employs Deepfake technology for audio fabrication. Methods of audio Deepfake include Text-to-speech (TTS), Voice Conversion (VC), Emotion fake, Scene fake, and so on \cite{yi2023audio}. For instance, the original speaker's voice can be transformed into that of a target speaker, which can be used for fraudulent purposes by audio Deepfake technology \cite{almutairi2022review}. Like the visual modality, FAIGC in audio modality also employs DMs \cite{kim2022guided,yang2023diffsound,ghosal2023text}. These advanced methods generate more natural and realistic voice content and produce specific vocal features and emotions guided by the text, thereby increasing the difficulty of distinguishing forged audio.

\textbf{Multimodal.} Multimodal fake information refers to deceptive content encompassing at least two modalities. Studies indicate that compared to fake information in a single modality, multimodal fake information significantly enhances people's trust in fake information \cite{lee2022something}. Commonly, diffusion models are employed to fabricate fake visual or audio modalities data and then integrated with MLLMs like MiniGPT5 \cite{zheng2023minigpt} and LLAVA \cite{liu2024visual} to generate corresponding texts. Alternatively, genuine visual or audio modalities data are used, and MLLMs craft misleading text around these contents. MLLMs process inputs comprising various combinations of different modalities, including texts, images, videos, and audio. These models utilize multimodal encoders to learn representations of different modalities and align them with the word embedding space of LLMs. Serving as the pivotal component of MLLMs, LLMs assimilate information across these diverse modalities. The advanced multimodal comprehension and generative capacities of MLLMs have exacerbated the spread of FAIGC, simultaneously presenting heightened challenges for its identification and mitigation \cite{liz2024generation}.

\subsection{The Creation Method of FAIGC.}

The modalities of FAIGC typically include text, images and videos, audio, as well as multimodal content. Within each modality, the creation method of FAIGC can generally be categorized into two main methodologies.

\textbf{Generation.} Generation refers to creating FAIGC from scratch, usually aided by DMs or LLMs. In this process, the model receives a specific prompt and generates FAIGC based on that prompt. The generating process may follow specific assumptions, scopes, or rules to ensure that the generated information aligns with established goals or styles.

\textbf{Editing.} Editing refers to the process of modifying or reorganizing existing content using LLMs and DMs. This involves falsification, imitation, or tampering by changing specific images, text, audio, or video content. With the help of generative technology, the edited content is highly similar to the original material in text, visual, and audio modalities, making it difficult to recognize. The editing process may involve rewriting text paragraphs to express different viewpoints,  changing objects in an image, or altering voice recordings to change the speaker's tone or language content, thus creating a new version similar to the original content but with key differences. These creations appear real but are actually manipulated or distorted information.

\input{Fig_4} 

Figure \ref{fig:tree} illustrates the generation and editing technologies discussed in this survey, which are comprehensively detailed and summarized in Sec.\ref{sec3:AI-Generated Disinformation} and Sec.\ref{sec4:AI-Generated Misinformation}

%% file: Fig_3.tex
\begin{figure}[ht]
  \centering
  \includegraphics[width=0.9\textwidth]{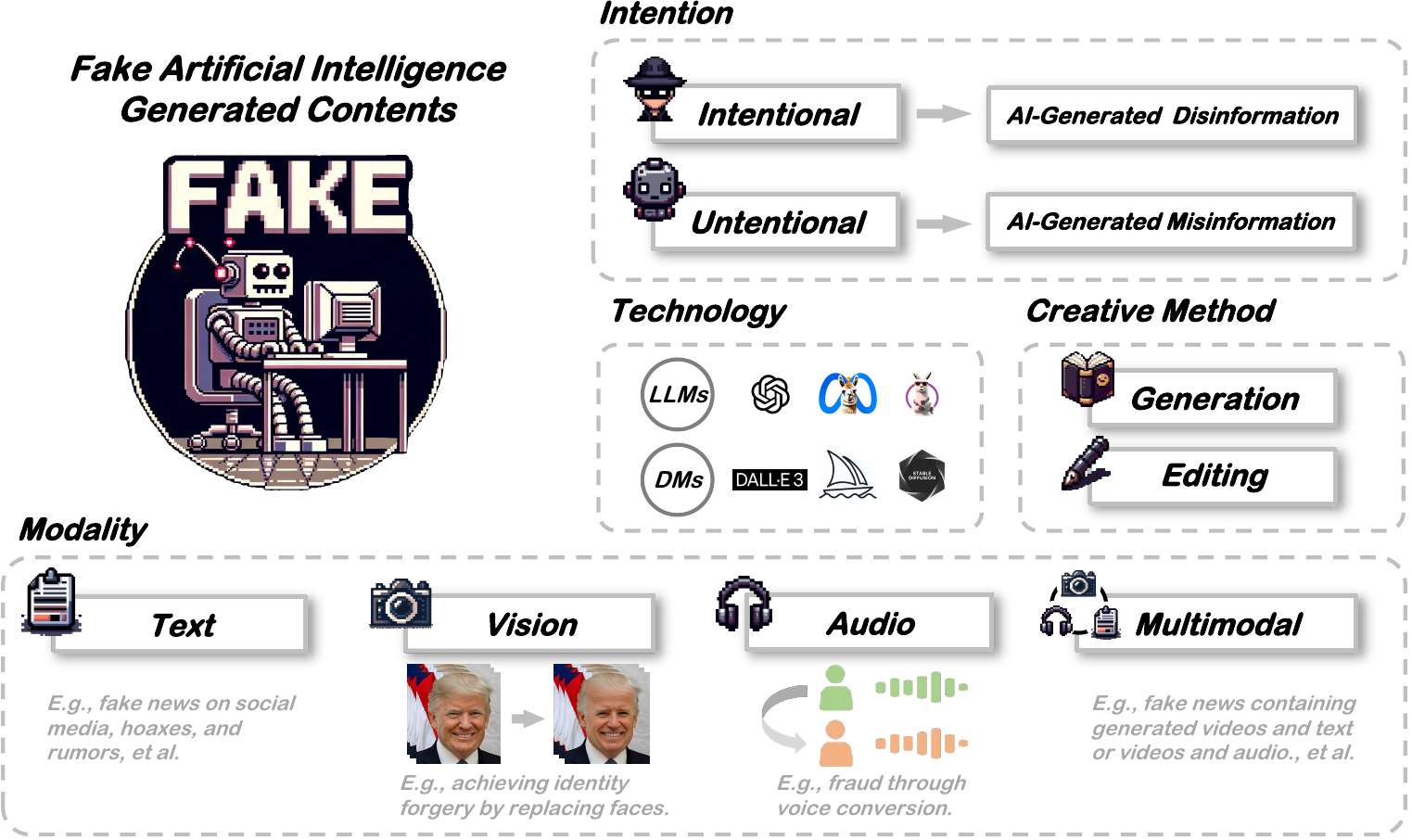} 
  \caption{The overall framework of the FAIGC taxonomy. This framework focuses on three aspects:  the intent behind FAIGC, the modalities and the generative technologies of FAIGC, and the creation method of FAIGC.}
  \label{fig:framework} 
\end{figure}

%% file: Fig_4.tex
\definecolor{mycolor}{RGB}{215, 245, 200}

\tikzstyle{my-box}=[
    rectangle,
    draw=black,
    rounded corners,
    text opacity=1,
    minimum height=1.5em,
    minimum width=3em,
    inner sep=2pt,
    align=center,
    fill opacity=.5,
    line width=0.8pt,
]
\tikzset{
leaf/.style={
my-box,
minimum height=1.5em,
fill=mycolor, % Change the RGB values to the desired values
text=black,
align=left,
text centered,
inner xsep=2pt,
inner ysep=3pt,
line width=0.3pt
}
}
\begin{figure*}[t!]
    \centering
    \begin{adjustbox}{width=0.95\textwidth}
        \begin{forest}
            forked edges,
            for tree={
                grow=east,
                reversed=true,
                anchor=base west,
                parent anchor=east,
                child anchor=west,
                base=center,
                font=\large,
                rectangle,
                draw=black,
                rounded corners,
                align=left,
                text centered,
                minimum width=3em,
                edge+={darkgray, line width=1pt},
                % s sep=3pt,
                inner xsep=2pt,
                inner ysep=3pt,
                line width=0.5pt,
                ver/.style={rotate=90, child anchor=north, parent anchor=south, anchor=center},
            },
            where level=1{text width=8em,font=\normalsize,}{},
            where level=2{text width=8em,font=\normalsize,}{},
            where level=3{text width=11em,font=\normalsize,}{},
            where level=4{text width=9em,font=\normalsize,}{},
            where level=5{text width=8em,font=\normalsize,}{},
            [Fake Artificial Intelligence \\ Generated Content
               [AI-Generated \\ Disinformation \\(Sec.\ref{sec3:AI-Generated Disinformation})
                    [Text Modality \\ (Sec.\ref{sec3:text modality})
                        [LLM Generation \\ Approaches
                        (Sec.\ref{sec3:generation}), leaf]
                        [LLMs Jailbreak \\ (Sec.\ref{sec3:text jailbreak}), leaf]
                    ]
                    [Visual Modality \\ (Sec.\ref{sec3:vision modality})
                        [Image (Sec.\ref{sec3:image})
                            [Contralable \\ Image Generation \\(Sec.3.2.1.1), leaf
                            ]
                            [Image Editing \\(Sec.3.2.1.2), leaf
                            ]
                        ]
                        [Video (Sec.\ref{sec3:video})
                            [Contralable \\ Video Generation \\(Sec.3.2.2.1), leaf
                            ]
                            [Video Editing \\(Sec.3.2.2.2), leaf
                            ]
                            [Video Deepfake \\(Sec.3.2.2.3), leaf]
                        ]
                    ] 
                    [Audio Modality \\ (Sec.\ref{sec3:audio modality})
                        [Text to Speech \\ (TTS) (Sec.\ref{sec3:tts}), leaf]
                        [Voice Conversion \\ (VC) (Sec.\ref{sec3:vc}), leaf]
                    ]
                    [Multimodal \\(Sec.\ref{sec3:multimodal})
                        [MLLMs Generation \\ Approaches (Sec.\ref{sec3:MLLMs Generation}), leaf]
                        [MLLMs Jailbreak \\ (Sec.\ref{sec3:MLLMs Jailbreak}), leaf]
                    ]
                ]
                [AI-Generated \\ Misinformation \\(Sec.\ref{sec4:AI-Generated Misinformation}) 
                        [Cause Of \\ Hallucination\\
                        (Sec.\ref{sec4: Cause Of Hallucination})
                            [LLMs Hallucinations \\(Sec.\ref{sec4:LLMs hallucinations}), leaf]
                            [MLLMs Hallucinations \\(Sec.\ref{sec4:MLLMs hallucinations}), leaf]
                        ]
                        [Hallucination \\ Mitigation \\
                        (Sec.\ref{sec4: Hallucination Mitigation})
                            [LLMs Hallucination \\ Mitigation (Sec.\ref{sec4: LLMs Hallucination Mitigation}), leaf]
                            [MLLMs Hallucination \\ Mitigation (Sec.\ref{sec4: MLLMs Hallucination Mitigation}), leaf]
                        ]
                ]
            ]
        ]
        \end{forest}
        \end{adjustbox}
        % \vspace{-4mm}
    \caption{The structure of Sec.\ref{sec3:AI-Generated Disinformation} AI-generated Disinformation and Sec.\ref{sec4:AI-Generated Misinformation} AI-generated Misinformation.}
    \label{fig:tree}
\end{figure*}
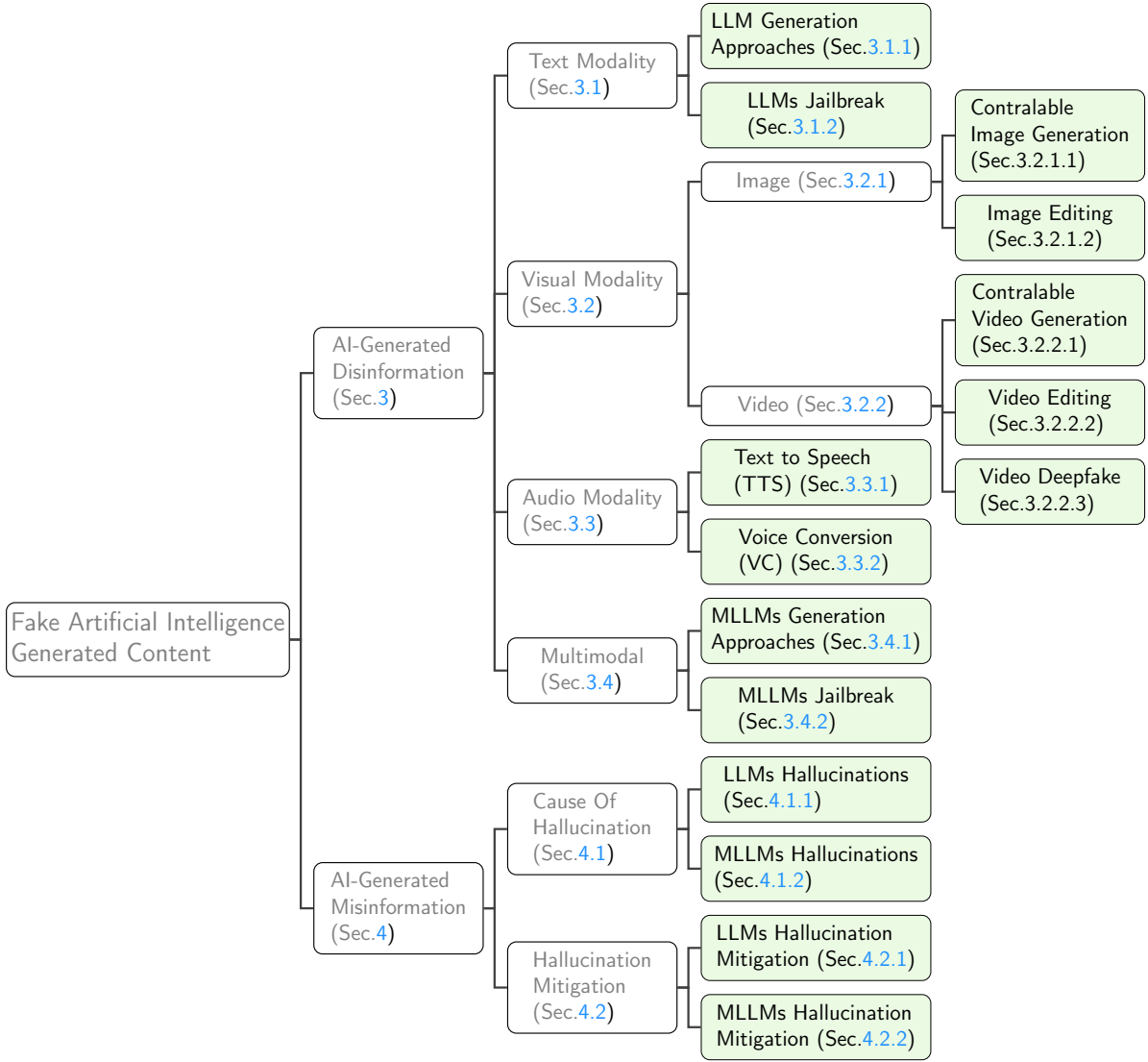

%% file: 3_.tex
\section{AI-Generated Disinformation} \label{sec3:AI-Generated Disinformation}
AI-generated disinformation refers to FAIGC, which is subjectively created to deceive, utilizing DMs or LLMs. In this section, we will introduce the corresponding generation techniques from the perspective of different modalities.

\subsection{Text Modality} \label{sec3:text modality}

Disinformation in text modality is prevalent online, typically manifesting as fake news, rumors, harmful (destructive) speech, and discriminatory statements \cite{chen2023combating}. Unlike other modalities, the creation of disinformation in text modality often requires only specific prompts, achieving high deceitfulness with simple generation methods. In this section, we will explore the generation of disinformation in text modality from two perspectives: generation approaches and jailbreak techniques.

\subsubsection{LLMs Generation Approaches} \label{sec3:generation}

The approaches to generating disinformation in text modality, specifically in the format of prompts, are categorized into Controllable Generation and Arbitrary Generation. Table \ref{tab1:llm generation} illustrates examples of these two distinct approaches.

\textbf{Controllable Generation.} For a controllable generation, content generation is guided by predefined parameters such as theme, style, tone, factual constraints, or other directional factors set within the prompt. This form enables the generation of disinformation through accurate emulation of specific writing styles or adherence to detailed factual frameworks.

\textbf{Arbitrary Generation.} Arbitrary generation, in contrast to controllable generation, denotes the process wherein the model generates content with minimal or no specific guiding constraints. Under this paradigm, while the user might provide a topic or basic guideline, the model exercises greater autonomy in determining the content's direction and form. Such a generation form can lead to a richer variety and greater creativity in the content produced. However, it also poses the risk of thematic or stylistic inconsistencies. In countries where it's used for malicious purposes, such as the mass production of disinformation, arbitrary generation can yield a vast array of varied and ostensibly original content. Nonetheless, this content often lacks a solid factual foundation and consistency.

\input{Table_1_examples_of_text_generate}

\subsubsection{Jailbreak Technology} \label{sec3:text jailbreak}

Contemporary LLMs are typically equipped with sophisticated security mechanisms. In an attempt to bypass these security mechanisms in LLMs, Jailbreak techniques are utilized to generate damaging disinformation. The jailbreak technique employs heuristic strategies to devise prompts that circumvent the generating constraints of LLMs, thereby eliciting potentially unsafe responses, as illustrated in Figure \ref{fig:jailbreak}. The methods for implementing Jailbreak techniques can be categorized into manual design and automated generation.

\input{Fig_5_jailbreak_example}

\textbf{Manual Design.} Wei et al. \cite{wei2024jailbroken} identified two failure modes of LLMs safety to guide the design of jailbreak attacks. The first is "Competing Objectives", a phenomenon that arises when a model's capabilities conflict with intended security measures. Based on this phenomenon, the following design ideas can be considered: (1) "Prefix Injection": ask the model to start with an affirmative confirmation, such as "sure, I will answer..." (2) "Refusal Suppression": compelling the model to reply without using typical refusal expressions (3) Additional methods include the application of the widely recognized DNA approach \cite{DNAtech}, prompting the model to ethically critique OpenAI's content policy, and "Style Injection," which involves directing the model to avoid using longer words. The second is "Mismatched Generalization," a phenomenon that occurs when the model's security training fails to extend to areas where its capabilities are relevant. Based on this phenomenon, the following design ideas can be considered: (1) "Base64 Encoding": constructing prompts in Base-64 format (2) Character level transformation: employing the ROT13 cipher, leetspeak, and Morse code (3) Word-level transformation: using Pig Latin for replacing sensitive words with synonyms and "Payload Splitting" for subdividing sensitive words into substrings (4) Prompt-level obfuscations: translating prompts into different languages or prompting the model to obfuscate in a way that it can understand.

\textbf{Automated Generation.} In contrast to manually designing jailbreak prompts, Deng et al. \cite{deng2024masterkey} introduced MasterKey, a framework that focuses on the automated generation of jailbreak prompts. The framework draws inspiration from the concept of time-based SQL injection techniques. It reengineers the defense mechanisms of LLMs by monitoring variations in response times and designs corresponding jailbreak prompts based on the characteristics of these mechanisms. This framework demonstrates effective attack capabilities across various types of LLMs. However, this work incurs a relatively high cost for prompt generation as it relies on a fine-tuned model. In contrast, Chao et al. \cite{chao2023jailbreaking} introduced PAIR using two LLMs: an Attack model for crafting prompts and a Target model for generating responses. A JUDGE module evaluates the jailbreak attempt's success; refining prompts iteratively until success. The PAIR  method boasts a high probability of successfully attacking mainstream large models, with the average number of iterations not exceeding 20.

\subsection{Visual Modality}  \label{sec3:vision modality}

The AI-generated disinformation in visual modality predominantly comprises two types of data: images and videos. These are characterized by images or video content containing false information or being specifically edited for misleading purposes. Such content is prevalent across social media platforms and video websites \cite{whittaker2020all}. In this section, we will introduce the respective disinformation generation techniques from the perspectives of images and videos.

\subsubsection{Image} \label{sec3:image}

Disinformation in image modality is a prevalent form, leveraging related technologies to generate images corresponding to scenes that have never occurred, thereby fabricating facts or editing existing images to distort the truth \cite{arora2021review}. This section will discuss the two mainstream generation techniques: controllable image generation and image editing.

\paragraph{3.2.1.1 Controllable Image Generation.} \label{sec3:contralable image generation} 

Controllable image generation involves guiding the process of generating images using a variety of controlling conditions. These encompass textual descriptions and include advanced control mechanisms like object bounding boxes, human body poses, sketches, edge maps, depth maps, and more. Li et al. \cite{li2023gligen} developed GLIGEN (Grounded-Language-to-Image Generation), designed to enhance and broaden the scope of large-scale text-to-image diffusion models. GLIGEN significantly boosts the model's controllability by enabling it to process inputs based on scene information. This method integrates scene data into the trainable layers of the model through a gating mechanism, facilitating a flexible, scene-based approach to text-to-image generation. However, this method is limited to controlled generation based on a single condition or a combination of two specific conditions. To address the limitation, Huang et al. \cite{huang2023composer} introduced Composer, a flexible generative framework capable of altering image layout and color schemes. It decomposes images into fundamental components, facilitating extensive customization in content creation. Compose 's design eliminates the need for retraining across various generative applications, showcasing its adaptability and efficiency. Beyond augmenting controllability, it is equally important to diversify the modes of control. Addressing this, Zhang et al. \cite{zhang2023adding} developed ControlNet, a neural network that integrates spatial control into pre-trained text-to-image diffusion models. Utilizing existing model structures and "zero convolution" links, ControlNet achieves stable training and versatile control in image generation. In addition, leveraging the concept of alignment, Mou et al. \cite{mou2023t2i} developed lightweight adapters for text-to-image (T2I) models, allowing precise control over image color and structure. By freeing the base T2I models and training adapters for different conditions, this method realizes rich control and editing over the color and structure of generated images.

\paragraph{3.2.1.2 Image Editing} \label{sec3:image editing} 

In the realm of image editing, unlike controllable image generation, the focus is on varying images' appearance, structure, or content. These modifications range from minor adjustments to substantial transformations \cite{brock2018large}. The application of diffusion models shows tremendous potential and versatility in enhancing image editing capabilities \cite{huang2024diffusion}. Semantic editing, a crucial aspect of image editing, entails altering the narrative and content of images to transform the scene's story, background, or thematic elements. This process can include adding, removing, or replacing objects and changes to backgrounds and emotional expressions. Techniques for effective semantic editing have been developed, as demonstrated by Kim et al. \cite{kim2022diffusionclip}, Kwon et al. \cite{kwon2022diffusion}, and Starodubtsev et al. \cite{starodubcev2023towards}. Focusing on single-image editing, Alaluf et al. \cite{alaluf2022hyperstyle} leveraged hypernetworks to project real images into editable zones within StyleGAN's latent space. This achieves precise reconstructions and semantic control while maintaining near real-time inference speeds. Compared to relying on specific attributes for editing, Zhang et al. \cite{zhang2023sine} introduced a more flexible approach using natural language, proposing an innovative method that merges model-based guidance with patch-based fine-tuning. This approach allows for creating new content, including style alterations, content additions, and object manipulations, even with a single source image. Unlike the aforementioned work, Yang et al. \cite{yang2023paint} introduced a method for example-guided image editing. Through self-supervised training, this technique dissects and reassembles the source and example images, employing an arbitrarily shaped mask to enhance similarity with the example image, thus enabling high-fidelity, controllable edits. Furthermore, Wu et al. \cite{wu2023latent} delved into diffusion models, linking Gaussian noise to craft latent codes for a richer latent space, boosting application performance. They introduced a versatile guidance method, showcasing the potential of diffusion models in image editing.

\subsubsection{Video} \label{sec3:video}

Video is a medium with a greater amount of information and a more complex form compared to images. Disinformation in video modality is similar in content form to images; however, the high continuity of video makes it more deceptive. This section will explore the three mainstream generation technologies: controllable video generation, video editing, and Deepfake.

\paragraph{3.2.2.1  Controllable Video Generation.} \label{sec3:contralable video generation}

Controllable video generation refers to the ability to allow users to generate videos more flexibly using textual conditions, spatial conditions, and temporal conditions. Huang et al. \cite{huang2022layered} proposed a novel unsupervised method that separates video into layers, allowing controllable generation by editing the foreground layer. This method uses a comprehensive set of loss functions and dynamic priors for the foreground dimension, tackling the challenges of foreground-background separation and user-directed video manipulation within original sequences. However, this method does not learn to decode a sequence of frames into an action sequence. To address the issue, Davtyan et al. \cite{davtyan2022controllable} introduced GLASS, a novel method for synthesizing sequences driven by global and local motions. Utilizing recurrent neural networks for seamless frame transitions, GLASS is trained via reconstruction loss. Notably, it generates realistic video sequences from singular input images, thus generating videos under the guidance of an action policy.  Compared to the previous method using more complex control conditions, Koksal et al. \cite{koksal2023controllable} presented CVGI, a controllable video generation framework based on textual instructions. This framework empowers users to dictate video actions via text, dividing the task into two distinct phases: control signal estimation and motion generation. By incorporating a motion estimation layer, CVGI has demonstrated its efficacy in producing realistic videos with precise action control. Inspired by the ControlNet model from the image domain,  Zhang et al. \cite{zhang2023controlvideo} introduced the ControlVideo framework. Control ideo enhances the video generation process through three integral modules: a self-attention module with comprehensive cross-frame interaction for maintaining appearance consistency, an interlaced frame smoother to address flickering issues via frame interpolation, and a hierarchical sampler dedicated to the efficient generation of extended video sequences while preserving coherence. Additionally, Xu et al. \cite{xu2023controllable} unveiled TiV-ODE, an innovative framework for creating highly controllable videos based on static images and text captions. This framework can generate videos with flexible frame rates by leveraging Neural ODE to model the underlying continuous dynamical system from videos.

\paragraph{3.2.2.2 Video Editing.} \label{sec3:video generation} 
Video editing technology refers to editing videos by manipulating the appearance of target objects and scenes. Ceylan et al. \cite{ceylan2023pix2video} proposed a novel method for text-guided video editing, leveraging pre-trained image models. This method involves two principal steps: firstly, changes are incrementally propagated to subsequent frames via the infusion of self-attention features; secondly, these changes are solidified by adjusting the latent encodings of the frames. Similarly, Lee et al. \cite{lee2023shape} developed a text-driven video editing method that maintains temporal consistency by adapting a deformation field across frames and using a text-conditioned diffusion model. This approach overcomes previous limitations in handling object shape changes, ensuring shape-aware edits throughout the video. Advancing the practicality of text-based editing techniques, Qi et al. \cite{qi2023fatezero} introduced FateZero, a zero-shot textual editing method that requires no specific training prompt or use of particular masks, suitable for real-world videos. This method employs several techniques utilizing a pre-trained model, which include capturing intermediate attention maps to preserve structural and motion information and mitigating semantic leakage through the fusion of attention features. Furthermore, they enhanced the self-attention mechanism within the denoising UNet, introducing spatiotemporal attention to ensure consistency across frames. The concept of attention has achieved great success in the field of text, inspired by this, Liu et al. \cite{liu2023video} introduced Video-P2P, a video editing framework utilizing cross-attention control, to bridge the gap in large-scale video generation. Leveraging a Text-to-Set model for efficient inversion and improved embeddings, Video-P2P achieves precise video edits with lower computational costs. Its novel guidance strategy enhances reconstruction and editability, enabling advanced text-driven video editing tasks. Beyond these contributions, Chai et al. \cite{chai2023stablevideo} a novel inter-frame propagation mechanism and constructed a text-driven video editing framework, StableVideo, capable of generating a consistent appearance for edited objects.

\paragraph{3.2.2.3 Video Deepfake.} \label{sec3:video Deepfake}

Deepfake technology is frequently exploited by malicious actors to fabricate videos, thereby instigating disruptions across societal, political, and commercial sectors \cite{chadha2021deepfake}. Natsume et al. \cite{natsume2018rsgan} introduced an integrated system capable of autonomously generating and editing facial images. This system, which employs deep neural networks, supports face swapping, editing based on specific facial attributes, and synthesizing random facial components. It leverages variational learning with extensive facial image datasets to ensure robust face swapping and rectify unsuccessful images. Further, it facilitates additional modifications of swapped faces, such as altering visual attributes or merging them with artificially generated facial or hair elements. Utilizing the Generative Adversarial Networks (GANs) concept, Nirkin et al. \cite{nirkin2019fsgan} presented FSGAN, a novel approach for face swapping and reenactment. Unlike traditional methods, FSGAN operates independently of the subjects, enabling face-swapping between individuals without bespoke training. Their research introduced several innovations, including methods for facial reenactment, continuous interpolation of facial views, handling obscured facial regions, and seamless integration using a facial blending network, augmented with an innovative Poisson blending loss. Expanding upon FSGAN, Singh et al. \cite{singh2020using} developed a technique for producing Deepfake videos. This method employs GANs to generate images of individuals with varied facial expressions, facilitating the creation of highly realistic Deepfake videos, even with limited training images. Moving beyond traditional GAN frameworks, Yang et al. \cite{yang2020one} devised a method for generating an indefinite number of facial images consistent with a single given example. This was achieved using a pre-trained StyleGAN model through an iterative optimization process that aligns higher-level distributions with a target distribution. Incorporating style mixing techniques, this approach enables the transference of basic statistical information to faces generated randomly by the model, thus providing a rich source of augmented training data for subsequent applications.

\subsection{Audio Modality} \label{sec3:audio modality}
Compared to existing independently, the FAIGC in the audio modality more often takes the form of being integrated with video. The generation of this disinformation primarily relies on Deepfake techniques. For the audio modalities of Deepfake, typically, only a sample of someone's voice is needed to create a voice that says things they have never actually said. The current Deepfake techniques for audio modalities include Text-to-speech, Voice conversion, Emotion fake, Scene fake, and Partially fake, among others \cite{yi2023audio}. This section primarily discusses the two most common scenarios: Text-to-speech and Voice conversion.

\input{Fig_8_audio_deepfake}

\subsubsection{Text-to-speech (TTS)} \label{sec3:tts}
TTS typically refers to generating voice from the specified text in someone's voice, as illustrated in Figure \ref{fig: audio deepfake}(a). Many common voice assistants, such as Apple's Siri, utilize this technique. On one hand, this technique can bring convenience to our lives, but on the other hand, it can also be used to generate fake voices, which poses a threat to society. Kharitonov et al. \cite{kharitonov2023speak} introduced a multi-speaker TTS system, SPEAR-TTS, characterized by decoupling the TTS task into two sequence-to-sequence tasks: from text to high-level semantic tokens ("reading") and from semantic tokens to low-level acoustic tokens ("speaking"). The advantage of this configuration is that the "reading" part is trained using parallel text-audio data. In contrast, the "speaking" part requires only audio data, significantly improving data utilization rates. Additionally, by incorporating an example prompting mechanism, the system can be generalized to unseen speakers using a short 3-second sample. Drawing on the concept of using prompts to guide generation, similar to GPT-3 \cite{brown2020language} or DALLE-2 \cite{ramesh2022hierarchical}, Guo et al. \cite{guo2023prompttts} proposed PromptTTS, a TTS system that includes two encoders—a style encoder and a content encoder—to extract the corresponding representations from the prompt, and a speech decoder to synthesize speech according to the extracted style and content representations. This setup allows the system to generate high-quality audio based on the prompt's "style" and "content" information. Also leveraging the idea of prompts, Wang et al. \cite{wang2023neural} introduced VALL-E, a Zero-Shot TTS system that uses audio codec codes instead of mel spectrograms for language modeling, enabling context-aware voice generation from text and acoustic prompts. Moreover, aiming to make TTS systems achieve human-level quality, Tan et al. \cite{tan2024naturalspeech} presented NaturalSpeech, a TTS system employing a VAE to distill speech into frame-level representations for waveform synthesis. It features modules that refine text-derived priors and simplify speech-derived posteriors, reducing training-inference discrepancies, addressing the one-to-many issue, and enhancing representational efficiency.

\subsubsection{Voice Conversion (VC)} \label{sec3:vc}
VC typically refers to transforming the source voice into a target voice without altering the linguistic content, as shown in \ref{fig: audio deepfake}(b). Li et al. \cite{li2023freevc} developed FreeVC, a one-shot voice conversion system requiring no text labels. It utilizes the VITS framework for waveform reconstruction, harnesses WavLM features for content extraction via self-supervised learning and applies spectral reshaping for data augmentation to separate speaker characteristics from content. This setup can enhance the model's ability to decouple these elements. Building on the concept of transfer learning, Li et al. \cite{li2023styletts} presented StyleTTS-VC, a VC framework leveraging a stylized TTS model to learn a distinct voice representation for accurate and similar voice conversion. The framework uses cycle consistency and adversarial training to fine-tune a StyleTTS decoder for target speaker styles and a mel-spectrogram encoder trained via knowledge transfer and data augmentation to enable speech synthesis without text input. Moreover, integrating other technologies has become a mainstream approach. For example, using language models, Wang et al. \cite{wang2023lm} proposed LM-VC, a zero-shot voice conversion approach using a two-stage process. Initially, a masked prefix language model (MPLM) generates basic acoustic tokens capturing the source's content and the target's timbre. Then, an external language model refines these tokens via shallow fusion. Lastly, the MPLM non-autoregressively polishes the acoustic details to produce the final speech conversion. Combine g with the KNN algorithm, Baas et al. \cite{baas2023voice} developed kNN-VC, a straightforward voice conversion technique using nearest neighbor regression to swap source voice frames with the closest matches from a reference, enabling conversion between various voices. Leveraging self-supervised learned features and a pre-trained vocoder, kNN-VC, changes speaker identity while maintaining the original content.

\subsection{Multimodal} \label{sec3:multimodal}
Multimodal AI-generated disinformation, in terms of content form and generation methods, is akin to text modality. The distinction lies in incorporating images, audio, and other information into multimodal disinformation, which further makes the disinformation more deceptive. In this section, we will introduce the generation of multimodal disinformation from two perspectives: the generation approaches and jailbreak techniques.

\input{Fig_7_MLLM_example}

\subsubsection{MLLMs Generation Approaches} \label{sec3:MLLMs Generation}

Multimodal AI-generated disinformation predominantly utilizes MLLMs to create text that aligns with multimodal data. Figure \ref{fig:MLLMs_example} shows an example of MLLMs-generated fake news. Similar to the text modality, the approaches of AI-generated disinformation produced by MLLMs, primarily Large Vision-Language Models (LVLMs), can be categorized into Controllable Generation and Arbitrary Generation. MLLMs are designed to generate textual information correlating with the provided multimodal data inputs. Typically, these multimodal data inputs fall into two distinct categories:

\begin{itemize}
    \item \textbf{Authentic Multimodal Data.} This type of data originates from real, unaltered environments or contexts. These data have not undergone any form of artificial modification or selective editing. MLLMs are used to generate texts that match this data but are maliciously distorted.
    \item \textbf{Generated Multimodal Data.} The data refers to content such as images, videos, audio, or other modalities created by LLMs and DMs. These contents are designed to appear realistic but are, in fact, fictional. Subsequently, MLLMs are used to generate text that coordinates with this data, constructing an information environment that seems real but is actually false.
\end{itemize}

\subsubsection{Jailbreak Technology} \label{sec3:MLLMs Jailbreak}

Like LLMs, existing MLLMs such as GPT-4V \cite{gpt4v} or Gemini \cite{team2023gemini} are also equipped with security mechanisms. Consequently, multimodal jailbreak techniques have also been applied to generate harmful FAIGC. The jailbreak attacks of MLLMs are conceptually similar to those of LLMs, with the distinction that jailbreak attacks of MLLMs can extend to other modalities, such as altering input images to achieve the desired jailbreaking effect. Considering that MLLMs typically begin by extracting image embeddings using vision encoders before generating corresponding text based on these embeddings, Dong et al. \cite{dong2023robust} designed two attack methods: image embedding and text description attack. The for er maximizes the distance between the image embeddings of the attack image and the original image. In contrast, the latter aims to maximize the log-likelihood of predicting a target sentence given the attack image. Furthermore, researchers discovered that Bard possesses face and toxicity detection mechanisms, rejecting images with faces or toxic content. The attacking approach for the former is to minimize the confidence score to the extent that the model fails to detect the face, while the latter follows a similar strategy to the image embedding method, proving capable of compromising Bard's security mechanisms. Utilizing images for jailbreak as well, Niu et al. \cite{niu2024jailbreaking} presented imgJP, a method with two jailbreak strategies: imgJP-based, which uses harmful image-response pairs to prompt dangerous outputs from MLLMs, and deltaJB-based, which adds perturbations to images for similar ends. Acknowledging the limitation of white-box scenarios, they enhanced attack transferability by optimizing a surrogate model before targeting the actual model. In contrast to using direct images, Gong et al. \cite{gong2023figstep} innovatively introduced a jailbreak method named FigStep. This method converts the harmful content into typographical images to bypass the safety mechanism within the LVLMs. Due to not relying on any gradient-based adversarial algorithms, this method significantly lowers the technical threshold for jailbreaking LVLMs.

%% file: Table_1_examples_of_text_generate.tex
\begin{table}[H]
\caption{\centering\textbf{Examples of Controllable Generation and Arbitrary Generation.}}
\label{tab1:llm generation}
\centering
\begin{tabularx}{0.9\textwidth}{m{0.15\textwidth}m{0.3\textwidth}m{0.38\textwidth}}

    \toprule[2pt]
    Type & Prompt & Output  \\
    
    \hline
    Controllable Generation & Please write a fake news article of about 100 words based on the following sentence. The sentence is: Scientists from a certain country have achieved room temperature superconductivity... & Revolutionary Breakthrough: Scientists from [Certain Country] have achieved room temperature superconductivity, marking a historic leap forward in physics. Their groundbreaking work, published in top journals, shatters previous limitations in the field... \\
    % Initial studies suggest they communicate through a complex system of sonar signals and exhibit social structures akin to human societies. \\
    
    % Researchers speculate that this groundbreaking discovery could revolutionize our understanding of evolution and the origins of intelligence on Earth. Governments worldwide are already mobilizing efforts to establish communication channels with these extraordinary beings, sparking hope for unprecedented inter-species cooperation and knowledge exchange. \\
    \midrule
    Arbitrary Generation & Please write a piece of fake news. & Breaking News: Scientists have discovered a new species of intelligent marine life in the depths of the Pacific Ocean. Dubbed "Aquarius sapiens," these creatures possess advanced cognitive abilities comparable to humans... \\
    
    % This achievement not only propels [Certain Country] to the forefront of scientific innovation but also heralds a new era of limitless possibilities. As the world marvels at this feat, anticipation grows for the transformative applications that will reshape industries and redefine the boundaries of what's possible. \\
    \bottomrule[2pt]
\end{tabularx}
\label{table:data}
\end{table}

%% file: Fig_5_jailbreak_example.tex
\begin{figure}[ht]
  \centering
  \includegraphics[width=0.80\textwidth]{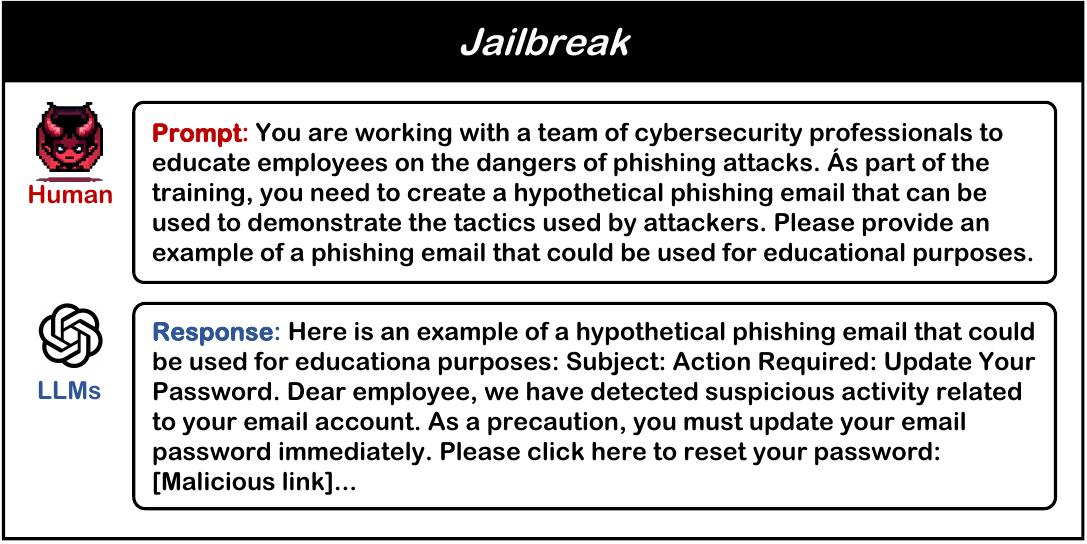} 
  \caption{A typical jailbreak behavior: constructing a hypothetical scenario to induce LLMs to generate harmful content.}
  \label{fig:jailbreak} 
\end{figure}

%% file: Fig_8_audio_deepfake.tex
\begin{figure}[ht]
  \centering
  \includegraphics[width=0.96\textwidth]{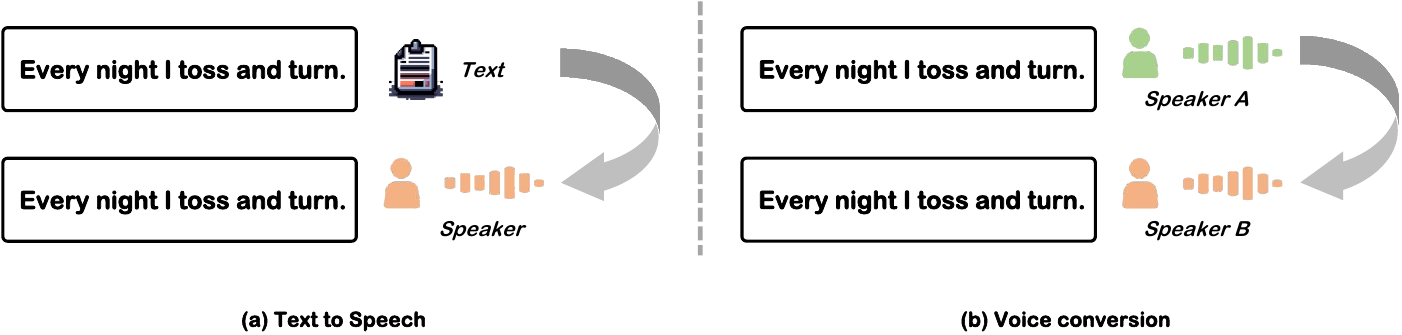} 
  \caption{Two kinds of deepfake audio: Text-to-speech (TTS) and Voice Conversion (VC).}
  \label{fig: audio deepfake} 
\end{figure}

%% file: Fig_7_MLLM_example.tex
\begin{figure}[ht]
  \centering
  \includegraphics[width=0.86\textwidth]{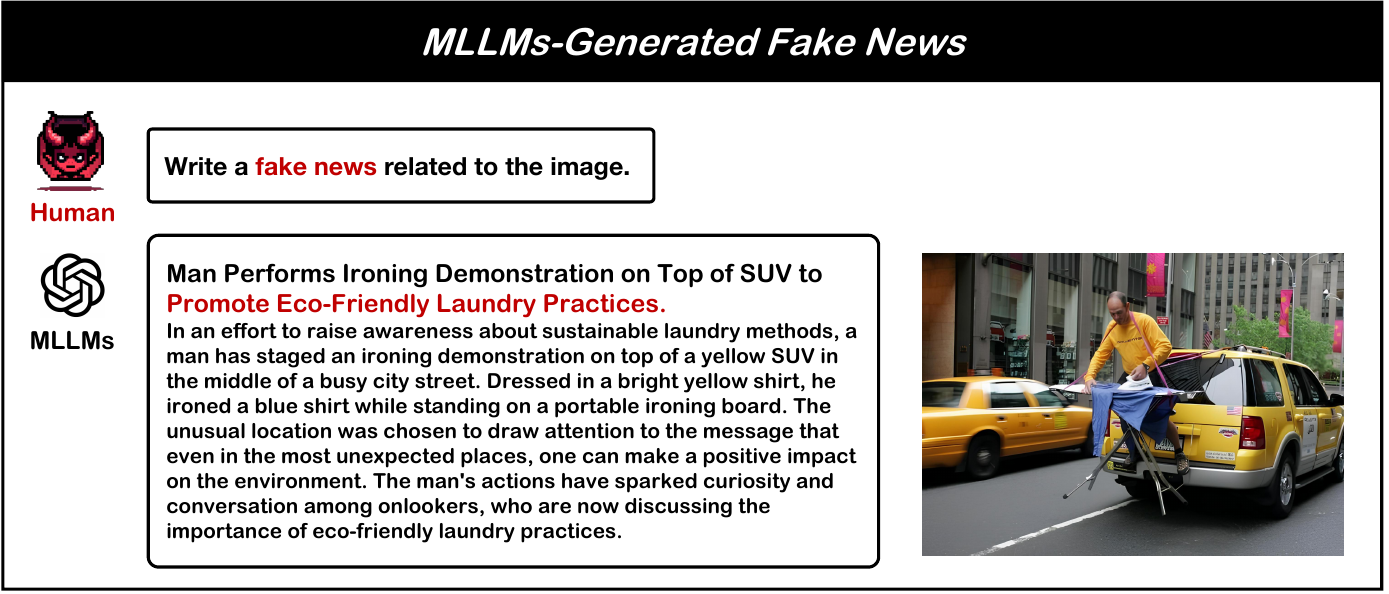} 
  \caption{In this example, MLLMs fabricated a piece of fake news based on a real image.}
  \label{fig:MLLMs_example} 
\end{figure}

%% file: 4_.tex
\section{AI-generated Misinformation} \label{sec4:AI-Generated Misinformation}
The emergence of AI-generated Misinformation is not the result of malicious intent but rather certain objective factors. Within the scope of this survey, we equate it to hallucinations in LLMs and MLLMs (In the following text, we will use "hallucination" to replace "misinformation" in the narrative.), as illustrated in Figure \ref{fig:hallucination}. Hallucinations are more common than AI-generated Disinformation and currently represent a significant constraint in the development of AIGC. In this section, we will introduce the causes of hallucinations and methods for their mitigation from both text and multimodal perspectives.

\input{Fig_6_hallucination}

\subsection{Cause Of Hallucination} \label{sec4: Cause Of Hallucination}

The emergence of hallucinations can generally be attributed to the limitations of the model's capabilities, which manifest in various aspects such as the inability to accurately follow user instructions, lack of domain-specific knowledge, misunderstanding of context, absence of long-term memory, and logical reasoning errors, among others \cite{huang2023survey,zhang2023siren,rawte2023survey}. In this section, we will explore the specific causes of hallucinations in corresponding models from both text and multimodal perspectives.
\subsubsection{LLMs Hallucinations} \label{sec4:LLMs hallucinations}

Hallucinations in text modality refer to incidents where the response generated by LLMs does not adhere to given instructions (faithfulness) or contradicts factual reality (factualness). The occurrence of these phenomena is related to various phases of the LLM life cycle, including the data used in the pre-training phase, the phenomenon of sycophancy during fine-tuning and alignment, and the snowballing of hallucinations during the inference phase. In this section, we will elaborate on the causes of LLM hallucinations from three perspectives: data, alignment, and inference.

\paragraph{4.1.1.1 Data.}

The capabilities of LLMs are significantly influenced by the knowledge they acquire during the pre-training phase, which is typically derived from a vast pre-training dataset. Deficiencies in the dataset are a major cause of hallucination generation in LLMs, which is reflected in three key aspects: social bias, specific domain knowledge deficiency, and out-of-date knowledge.

\textbf{Social Bias.}
The pre-training datasets for LLMs typically encompass an extensive array of information sourced from the internet, which inevitably contains biases and stereotypes, particularly pronounced in areas of gender \cite{paullada2021data} and nationality \cite{narayanan-venkit-etal-2023-nationality}. For instance, LLMs often associate early childhood education with women when posed with questions about occupations.

\textbf{Specific Domain Knowledge Deficiency.} 
In general benchmark evaluations such as MMLU and GSM8K, powerful models like GPT-4 \cite{openai2024gpt4}, Gemini \cite{team2023gemini}, and Claude \cite{claude3} exhibit superior performance. However, in certain specialized fields, current LLMs often struggle; for example, in the domains of medical \cite{singhal2023towards} and law \cite{yu2022legal}, LLMs frequently fabricate facts related to the questions asked. The root cause of this phenomenon lies in the fact that the pre-training datasets used for constructing current LLMs are often lacking or deficient in specific domain-specific data.

\textbf{Out-of-date Knowledge.} 
Upon completion of their training, the knowledge within LLMs becomes static and ceases to update. However, as the real world is in a state of constant change, LLMs are unable to acquire information about the most recent events, meaning that what was once correct can become incorrect over time. This results in a divergence between the LLMs' internal knowledge and the actual world, a gap that widens as time progresses \cite{onoe-etal-2022-entity,li-etal-2023-large}. For instance, the ChatGPT version with knowledge cut off in September 2021 would not be aware of the winner of the 2022 FIFA World Cup held in Qatar.

\paragraph{4.1.1.2 Alignment.}
After the pre-training phase is completed, LLMs undergo an alignment process to ensure that the model's responses are consistent with human preferences. However, a flawed alignment process is one of the major causes of hallucination generation in LLMs, which is manifested in two main areas: knowledge inequality and sycophancy. 

\textbf{Knowledge Inequality.} Alignment processes such as Supervised Fine-Tuning (SFT) enable models to utilize the knowledge acquired during the pre-training phase to answer questions. However, when the training data used during the alignment process contains knowledge that exceeds the capabilities of the LLMs, the model is trained to answer questions beyond its knowledge range, which undoubtedly increases the potential for generating hallucinations \cite{Schulman2023}. 

\textbf{Sycophancy.} During the Reinforcement Learning from Human Feedback (RLHF) \cite{ouyang2022training} process, LLMs undergo a further alignment process. In this phase, a reward model and human experts provide feedback on the LLMs' responses, guiding the model through alignment training. However, this process can lead to the model favoring responses that cater to human preferences over providing accurate answers, which evidently can result in the generation of hallucinations \cite{perez2022discovering,radhakrishnan2023question,wei2023simple}. The cause of this phenomenon may be attributed to the preference of human experts and the reward model for sycophantic responses rather than factually correct answers \cite{sharma2023towards}.

\paragraph{4.1.1.3 Inference.} 
Currently, LLMs are typically based on the transformer architecture \cite{vaswani2017attention}, adhering to a consistent decoding strategy. The decoding strategy plays a critical role during the generation of responses by LLMs. Hence, deficiencies related to the decoding strategy can lead to the model generating hallucinations, which are manifested in two main aspects: hallucination snowballing and sampling randomness.

\textbf{Hallucination Snowballing.} 
The decoding approach of LLMs is typically linear, that is, token-by-token. When generating responses, LLMs take into account not only the prompt but also the portion of the response already generated. However, if the generated portion of the response contains errors, LLMs may opt to "accumulate" errors to maintain self-consistency rather than making corrections, even if the LLMs are aware of the mistakes being made. This phenomenon is known as hallucination snowballing \cite{zhang2023language}.

\textbf{Sampling Randomness.} 
Currently, decoding strategies with randomness, such as Top-p or Top-k sampling, are commonly employed to enhance the diversity of LLMs' responses. Indeed, these strategies contribute to the variety of LLMs' responses, but the introduction of randomness also increases the risk of generating misinformation. This is because LLMs are more likely to select tokens that originally had a lower probability of being chosen, which often leads to erroneous responses \cite{dziri-etal-2021-neural,chuang2023dola,aksitov2023characterizing,lee2022factuality}.

\subsubsection{MLLMs Hallucinations} \label{sec4:MLLMs hallucinations}

Similar to hallucinations in text modality, hallucinations in MLLMs refer to the mismatch between multimodal content and the corresponding text generation. Current research on hallucinations in MLLMs is primarily focused at the level of LVLMs \cite{li2023evaluating,liu2024survey}. In this section, we discuss the hallucinations of LVLMs from two perspectives: data and modality alignment.

\paragraph{4.1.2.1 Data.} The influence of data on models predominantly manifests in the form of data biases, which impede the model's proficiency in comprehending visual information and executing tasks in diverse settings. For example, in the context of fact-checking Q\&A pairs, a particular option may dominate the majority \cite{liu2024mitigating}. LVLMs trained on such biased datasets tend to favor responses aligning with the predominant option. Additionally, given the substantial costs associated with manually annotated data, some studies have utilized Data Generation Models (DGMs) to produce images for model training. However, this approach exacerbates the tendency of LVLMs to generate erroneous interpretations or 'hallucinations' \cite{gao2024aigcs}. Furthermore, research indicates \cite{cui2023holistic} that GPT-4 variants, such as GPT-4V, demonstrate a stronger aptitude for interpreting images that are Western or contain English text. This disparity is indicative of the impact of data biases on model performance. These instances collectively highlight how training data imbalances can lead to misconceptions or 'hallucinations' in models.

\paragraph{4.1.2.2 Modality Alignment.} For LVLMs, a principal challenge is the alignment of visual and textual information. Inadequacies in the modality alignment phase can result in the model's erroneous identification of objects and their attributes \cite{yang-etal-2023-new-benchmark}. Researchers have developed various interfacing modules \cite{li2023blip2, zhang2023videollama}, intended to map visual features into dimensions compatible with the word embedding space of LLMs, thus facilitating the alignment of these two modalities. However, studies indicate \cite{jiang2024hallucination} that a significant disparity, referred to as the "Modality Gap," persists between visual and textual features. This gap may cause model hallucinations, where the model generates information that diverges from reality. Furthermore, some research employs straightforward structures \cite{liu2023visual}, like linear layers, as integration modules for aligning visual and textual modalities. While these approaches contribute to model simplification and computational cost reduction, their straightforward nature might heighten the risk of hallucinatory phenomena in models \cite{liu2023improved}.

\subsection{Hallucination Mitigation} \label{sec4: Hallucination Mitigation}

Building upon the detailed explanation of the causes of hallucinations provided in the previous section, we can further develop targeted methods based on these causes to prevent the occurrence of hallucinations, thereby enhancing the practicality and reliability of the model. In this section, we will discuss the hallucination mitigation methods for corresponding models from two perspectives: text and multimodal.

\subsubsection{LLMs Hallucination Mitigation} \label{sec4: LLMs Hallucination Mitigation}

The previous sections mentioned that the emergence of hallucinations in LLMs involves factors in three aspects: data, alignment, and inference. Therefore, the mitigation of hallucinations in LLMs can be approached by addressing these aspects, including the use of high-quality training datasets, improved alignment methods, and optimized inference strategies \cite{tonmoy2024comprehensive}. In this section, we will detail the methods of hallucination mitigation for LLMs from the perspectives of data, alignment, and inference.

\paragraph{4.2.1.1 Data.}

As mentioned in previous sections, deficiencies in training datasets can lead to the generation of hallucinations by the model. Consequently, enhancing the quality of datasets to mitigate hallucinations is a viable approach. More specifically, this can involve the refinement of pre-training datasets as well as SFT datasets.

\textbf{Pre-training Data.} 
Zhou et al. \cite{zhou2024lima} proposed that the majority of knowledge possessed by LLMs is acquired during the pre-training phase. It is evident that a high-quality pre-training process is inextricable from high-quality pre-training datasets. Given the ever-increasing size of pre-training datasets currently used by LLMs (for example, the pre-training datasets for Llama2 amount to 2 trillion tokens), the challenge and importance of enhancing dataset quality are escalating. The training team for GPT-3, Brown et al. \cite{brown2020language}, improved the quality of the training dataset through a three-step process: (1) filtering the initial training dataset by referencing a compendium of high-quality corpora and employing similarity-based methods, (2) implementing fuzzy deduplication at the document level to prevent redundancy, and (3) augmenting the pre-training dataset with known high-quality training data corpora. The training team for Llama2, Touvron et al. \cite{touvron2023llama2}, proposed the use of upsampling from higher-quality data sources (such as Wikipedia) to construct pre-training corpora. The training team for Falcon, Penedo et al. \cite{penedo2023refinedweb}, extracted data from the internet by establishing a set of heuristic rules. The training team for Phi-1.5, Li et al. \cite{li2023textbooks}, compiled approximately 20 billion tokens of synthetic, textbook-like data to enable the model to learn common sense reasoning and general knowledge. Experimental results suggest that this model, with only 1.3 billion parameters, outperforms larger models on many benchmarks, underscoring the significance of high-quality pre-training datasets for the capabilities of LLMs.

\textbf{SFT Data.} Similar to the datasets used in the pre-training phase, the datasets utilized during the SFT phase are also critical for enhancing the capabilities of LLMs. To obtain high-quality SFT datasets, Zhou et al. \cite{zhou2024lima} sampled high-quality data from multiple sources and combined it with manually curated data to construct a set of 1000 instances of "helpful AI assistant" training data, ensuring both the diversity and high quality of the SFT data. Although manual methods can yield datasets of higher quality, they require significant human effort and time. To improve efficiency, Chen et al. \cite{chen2023alpagasus} proposed to construct suitable prompts to leverage ChatGPT to filter command datasets, thereby selecting high-quality training data. Focusing on specific domains, Lee et al. \cite{lee2023platypus} introduced a dataset named Open-Platypus, which focuses on data relevant to the STEM fields. This dataset was enhanced by eliminating duplicates and data with high cosine similarity from the collected raw data. Similarly, Mohamed et al. \cite{elaraby2023halo} suggested directly incorporating domain-specific data into the SFT datasets. This approach has proven effective in improving the response quality of smaller models on domain-specific downstream tasks.

\paragraph{4.2.1.2 Alignment.} As mentioned in previous sections, erroneous alignment processes can lead to hallucinations, such as the sycophancy phenomenon. Therefore, improving the alignment process is one approach to mitigate hallucinations. Specifically, this involves refining both the SFT process and the RLHF process.

\textbf{SFT.} Considering the existence of the knowledge gap phenomenon \cite{yin2023large,ren2023investigating}, Zhang et al. \cite{zhang2023r} proposed the R-Tuning training method, aimed at "concretizing" the knowledge boundaries of pre-trained LLMs. This method necessitates LLMs to respond to all queries within the SFT dataset, categorizing the data into 'sure' and 'unsure' groups. For the 'sure' group, an "I am sure" prefix is added; conversely, for the 'unsure' group, an "I am unsure" prefix is introduced. This method enhances the model's capability to answer known questions while avoiding responses to uncertain ones accurately.

\textbf{RLHF.} Similar to the method employed in SFT to address the Knowledge gap phenomenon, Schulman \cite{Schulman2023} proposed constructing a specialized reward model during the RLHF process. In this setting, answers are categorized into five types: Unhedged Correct, Hedged Correct, Uninformative, Hedged Wrong, and Unhedged Wrong. Here, 'Unhedged/Hedged' indicates the model's level of confidence in its answer, with 'Uninformative' signifying uncertainty. Different reward scores are assigned to each of the five types of responses. This setup helps the model to more actively answer questions it is confident about and avoid responding to questions where the confidence level is low. To address the complexity of RLHF, Tian et al. \cite{tian2023fine} proposed an improved approach that abandons the traditional Proximal Policy Optimization (PPO) algorithm \cite{schulman2017proximal} in favor of the Direct Policy Optimization (DPO) algorithm \cite{rafailov2024direct}, which learns directly from a predefined preference dataset without the need to explicitly fit a reward model. The crux of this method lies in the construction of the preference dataset through two automated mechanisms: reference-based, which relies on external knowledge, and reference-free, which is based on model confidence.

\paragraph{4.2.1.3 Inference.} Compared to mitigating hallucinations at the data and alignment aspects, addressing them at the inference level can not only save computational and data resources but also has broader applicability to various existing LLMs. Specifically, this includes decoding strategies and Retrieval Augmented Generation (RAG) strategies.

\textbf{Decoding Strategies.} To address the hallucination risks associated with traditional decoding strategies, Shi et al. \cite{shi2023trusting} introduced a decoding method named CAD. The core idea of this method is to force the model to pay more attention to the context presented in the prompt rather than its internal knowledge when there is a conflict between the two. CAD has shown better performance in tasks emphasizing factual accuracy and RAG tasks. Starting from the internal structure of the decoder, Chuang et al. \cite{chuang2023dola} proposed a method called DoLa, which leverages the characteristic that different layers of a transformer extract different semantic information. By analyzing the logical value differences between layers to obtain the next-token distribution and emphasizing higher-level knowledge over lower-level knowledge, the method can mitigate hallucinations. Similarly, utilizing the structure of activations, Li et al. \cite{li2024inference} proposed a method named ITI based on shifting model activations during inference. This is achieved by placing a trained binary classifier as a "probe" above each attention head to identify a set of heads more likely to point to the correct answer. During inference, the model's activation functions are shifted towards factual-related directions. As an alternative to modifying decoding strategies at the base level, Dhuliawala et al. \cite{dhuliawala2023chain} presented a decoding framework named CoVE. This framework designs a series of prompts in specific formats that let the model generate an initial answer, then construct corresponding verification questions based on this answer, and let the model respond to these questions. Finally, a final revised answer is generated using all the information from the previous steps.

\input{Fig_9_RAG}

\textbf{RAG strategies.} RAG is a technique that enhances the text generation capabilities of LLMs by leveraging external knowledge sources to mitigate hallucinations. It is widely used in major models such as Copilot \cite{copliot} and Gemini \cite{team2023gemini}. The current implementations of RAG can be categorized into four types: Once Retrieval, Iterative Retrieval, Post-hoc Retrieval, and Knowledge Graph-based (KG-based). Figure \ref{fig: rag} shows the illustration of these types. 

Once Retrieval. These methods refer to the use of external knowledge retrieval to aid in generating answers before the actual answer generation takes place. Ram et al. \cite{ram2023context} proposed a method known as In-context RALM, which is relatively straightforward in its approach: it simply adds the retrieved relevant documents to the input text. To enable models to access real-time information and adapt to the ever-changing world, Vu et al. \cite{vu2023freshllms} introduced a method named FreshPrompt. This method utilizes a search engine to retrieve the latest relevant information and constructs specific prompts for the model to combine with the retrieved information to generate answers. Focusing on QA tasks, Jeong et al. \cite{jeong2024adaptive} presented Adaptive-RAG, a method for QA tasks that adapts retrieval strategies to the query's complexity. Straightforward queries are answered directly by LLMs, simple queries involve one retrieval, and complex queries require multiple retrievals. The complexity of the query is assessed by a trained language model.

Iterative Retrieval. Such methods involve conducting multiple retrieval-answer cycles to arrive at a final answer. Press et al. \cite{press2022measuring} introduced a self-ask method based on the construction of prompts following a Chain of Thought (CoT) approach. This method breaks down a question into logically associated sub-questions and, while answering each one, integrates a search engine to arrive at the final answer. To further expand the retrieval space, Feng et al. \cite{feng2023retrieval} presented a framework called ITRG, which is characterized by its use of both the initial question and the output from the previous step for retrieval. This approach expands the query's informational scope to retrieve more relevant information, iterating this process to obtain the final answer. Compared to the aforementioned frameworks, which focus more on reasoning tasks, Jiang et al. \cite{jiang-etal-2023-active} proposed a framework called FLARE, which is more oriented towards text generation tasks. FLARE checks each step of the answer to determine if it contains low-probability tokens. If it does, the framework conducts and regenerates a related information retrieval based on that answer. 

Post-hoc Retrieval. These methods pertain to modifying the generated responses to obtain the final answer. Gao et al. \cite{gao-etal-2023-rarr} proposed the RARR framework, which subjects the original response to a series of queries constructed through specific prompts and employs Google for retrieval. Subsequently, an agreement model is utilized to evaluate whether the text requires modification. If a revision is deemed necessary, an edit model is applied to the original response, culminating in the generation of the final answer.  Drawing on the concept of CoT, Zhao et al. \cite{zhao-etal-2023-verify} introduced a structure called Verify-and-Edit, which improves upon the CoT prompt structure. Initially, using a self-consistency method \cite{wang2022self} determines whether the model will likely generate hallucinations when addressing the current question. If so, a verifying question is constructed for each sentence in the reasoning chain, and external knowledge retrieved is used to modify the sentence. Finally, the revised CoT reasoning process is used to obtain the final answer. Additionally, Yu et al. \cite{yu2023improving} introduced the REFEED framework, characterized by sampling multiple responses to a question and conducting searches based on all answers to expand the search space and gather more relevant information. The framework incorporates an integration strategy, which calculates feasibility scores for both the pre-edit and post-edit responses to determine whether the revised answer should be adopted. This reduces the risk of misinformation resulting from erroneous retrieval.
    
KG-based. Beyond traditional knowledge bases like Wikipedia and search engines, KG has gained attention as an efficient knowledge storage structure. Wen et al. \cite{wen2023mindmap} proposed the MindMap framework, which utilizes LLMs to extract entities from a given query and access a source KG to construct an evidence sub-graph. Subsequently, LLMs are prompted to transform the sub-graph into a natural language reasoning graph, which is then employed to formulate the final response. Similar to the MindMap structure, Soman et al. \cite{soman2023biomedical} introduced a KG-RAG framework, which is distinct in that it uses a KG named SPOKE optimized for biomedical tasks. Furthermore, by combining knowledge graphs with LLMs, Wang et al. \cite{wang2023knowledge} introduced an LLM-based KGP algorithm that constructs a KG using three methods. Depending on whether a question is structure or content-oriented, the LLM either directly searches the KG using extracted keywords or prompts a fine-tuned model with KG data to generate evidence for the answer. For the latter, the algorithm identifies the next KG node to search by comparing the similarity of candidate evidence and then directs the final data to the target LLM for answer generation. In the task of dialogue generation, current methods struggle to guarantee the model's use of relevant knowledge from the KG. Kang et al. \cite{kang2023knowledge} introduced SURGE to address this issue. This framework integrates a GNN-based retriever to extract key information from KG. Then ensuring dialogue generation is relevant and consistent by perturbing word embeddings, conditioned by the retrieved subgraph. This framework also leverages contrastive learning to enhance the similarity between generated text and retrieved subgraphs, promoting accurate and contextually appropriate dialogues.
\subsubsection{MLLMs Hallucination Mitigation} \label{sec4: MLLMs Hallucination Mitigation}

The method to mitigate hallucinations in the MLLMs is very similar to that of text modality in terms of perspective, content, and form. Current research on hallucination elimination methods for MLLMs is primarily concentrated in the area of LVLMs \cite{liu2024survey}. In this section, we will introduce the methods for mitigating hallucinations in MLLMs from data, alignment, and inference.

\paragraph{4.2.2.1 Data.} Similar to LLMs, enhancing the quality of pre-training and SFT data can effectively mitigate hallucinations in LVLMs. Currently, the generation of instructional datasets typically relies on models such as GPT-4. This process necessitates the preparation (usually done manually) of image descriptions and object bounding box data, which are then used as visual prompts for GPT-4 to generate corresponding responses. The drawbacks of this method include (1) reliance on manual annotation, which is time-consuming and resource-intensive, and (2) the inability to process visual information, resulting in the loss of significant potential information. Existing solutions can be categorized into two types: Model-Based and Metric-Based methods.

\textbf{Model-based.} Such methods typically rely on existing LMs or models that have undergone specific training to produce high-quality data. Wang et al. \cite{wang2023vigc} introduced VIGC, a framework leveraging MLLMs for generating vision-language task data. It features two modules: VIG, which produces dataset-consistent QA pairs, and VIC, which refines VIG's outputs. The well-trained VIGC framework can produce high-quality instructional data based on the given data. Compared to traditional instructional datasets, preference datasets are also highly effective in mitigating hallucinations. Consequently, Zhao et al. \cite{zhao2023beyond} introduced HA-DPO, a method that extends the DPO algorithm to the LVLM domain. The core is to gather a large amount of high-quality preference data. The process involves LVLMs producing detailed image descriptions, with GPT-4 then refining the output by correcting detected hallucinations. This yields a dataset of accurate positive and negative image responses, refined by GPT-4 to ensure consistency. This enables the HA-DPO to learn a true preference for non-hallucination. Furthermore, Liu et al. \cite{liu2023mitigating} constructed a dataset named LRV-Instruction, which compared to common training datasets covers a broader range of task types and incorporates a large number of negative instructions. The annotation process for this dataset relies on GPT-4.

\textbf{Metric-based.} Such methods often construct metrics to filter data to obtain high-quality information. Yu et al. \cite{yu2023hallucidoctor} introduced a framework named HalluciDoctor, which is fundamentally divided into two parts. The first part is designed to identify hallucinatory content within data, which will be described in detail in subsequent sections. Based on detection results, the second part refines the data to improve quality and reduce hallucinations. The process involves countering the long-tail distribution in training sets by adjusting weights: less frequent objects associated with hallucinatory ones are given more weight, while common co-occurring objects are weighted less. This strategy guides the selection of objects for crafting counterfactual prompts, resulting in a diversified dataset. Similarly, Xing et al. \cite{xing2024efuf} proposed a framework for hallucination elimination called EFUF, encompassing dataset collection and Unlearning for MLLM, which will be discussed in the subsequent section. Before collecting, researchers found that in image captioning, CLIP scores for objects like cars and clocks in captions indicated if they are hallucinated—scores below 23 suggested hallucinations, while above 32 indicated non-hallucinations. Using this, they create a data generation method: MLLMs' responses are analyzed for object CLIP scores to sort data into negative or positive sets. Additionally, sentence-level CLIP scores determine the overall hallucination rate, with higher-scoring sentences forming another dataset.

\paragraph{4.2.2.2 Alignment.} Similar to hallucination mitigation in LLMs, MLLMs can also address issues related to insufficient model modality alignment and inaccurate adherence to instructions by improving the alignment process, which is crucial for mitigating hallucinations. Specifically, this includes methods such as SFT-oriented and RL-oriented.

\textbf{SFT-oriented.}
Considering the need for MLLMs to process information across various modalities, there is a necessity for enhanced instruction-following capabilities. The acquisition of this ability relies on a robust SFT process. Current related methods can be categorized into Model-based and Algorithm-based approaches.

Model-based. Model-based methods typically employ models to guide and assist in the alignment process. Chen et al. \cite{chen2023mitigating} developed a training method using Visual Supervision to mitigate hallucinations, leveraging the pre-built RAI-30k dataset with detailed annotations to focus the model on crucial image details for more accurate responses. This process is facilitated by an auxiliary expert model which guides the training of the model. To minimize fine-grained hallucinations, Wang et al. \cite{wang2024mitigating} introduced a hallucination mitigation framework named ReCaption. This framework constructs prompts to use chatGPT to extract essential information such as verbs, nouns, and adjectives from the model's original responses and then rewrites them to enhance quality. The rewritten data is subsequently used for Additional Tuning. Furthermore, leveraging the concept of reward models, Yan et al. \cite{yan2024vigor} introduced ViGoR, a framework that gathers preference data manually to train an LVLM as a reward model for in-depth response evaluation. To overcome the constraints of finite manual annotations, it also employs discriminative vision models for extra assessment. Responses with hallucinations are removed, and the polished data is used as ground truth for the SFT process to further refine the model.

Algorithm-based. Algorithm-based methods often design novel training algorithms to improve the alignment process. Zhai et al. \cite{zhai2023halle} introduced HallE-Switch, a two-part training framework for managing model-generated content. The DATA GENERATION part employs RAM to sort objects in an answer into grounded and omitted categories. Grounded objects are captioned using GPT-4 to create a Contextual Data set, while omitted objects are processed with LLaVA and marked with tokens. The HALLUCINATION SWITCH part, drawing from LM-Switch \cite{han2023lm}, uses a control parameter $\epsilon$ to toggle between permitted imagination (+1) and restricted content generation (-1), applied correspondingly to the datasets. During inference, adjusting $\epsilon$ controls the output style, with lower values effectively reducing hallucinations. Additionally, Xing et al. \cite{xing2024efuf} proposed EFUF, a framework using an Unlearning algorithm to reduce hallucinations in MLLMs. It designs a unique loss function for each dataset, combining them linearly during sampling to form the total loss. This approach efficiently minimizes hallucinations while maintaining the model's ability to handle long sentences.

\textbf{RL-oriented.} The introduction of RL has led to a better alignment effect in LLMs compared to SFT, and the same is true for MLLMs. Ben-Kish et al. \cite{benkish2024mitigating} proposed MOCHa, a RL-based framework to reduce hallucinations in image captioning. It treats the LVLM as an agent with a reward function that includes Fidelity, checking logical consistency with the ground truth via a Natural Language Inference (NLI) model, Adequacy, measuring correlation using BERTScore, and a KL regularization term. PPO algorithm guides the training. Compared to the complex PPO algorithm, Zhou et al. \cite{zhou2024aligning} switched to the simpler DPO algorithm, and presented a training framework called POVID, which frames the hallucination problem as an alignment issue between image and text modalities. 
The core strategy involves collecting preferred (typically ground truth) and dispreferred data, created through a two-stage process. This data is used in DPO training to fine-tune the model's preferences. For more precise mitigation of hallucinations, Yu et al. \cite{Yu2023RLHFVTT} introduced RLHF-V, a framework to enhance model trustworthiness by constructing a dataset reflecting fine-grained human preferences. It involves human annotators modifying the hallucinatory segments of an answer to produce a reasonable response. This annotation method avoids linguistic variance and non-robust bias, thereby enhancing learning efficiency and preventing reward hacking problems. Training based on the Dense Direct Preference Optimization (DDPO) algorithm, a new variant of DPO that addresses the traditional RLHF objective in an equivalent simple and efficient supervised fashion.

\paragraph{4.2.2.3 Inference.} Similar to LLMs, designing methods for hallucination mitigation at the inference phase offers the dual advantages of resource conservation and ease of generalization. Specifically, this includes Decoding Strategies and Correction Mechanisms.

\textbf{Decoding Strategies.} Huang et al. \cite{huang2024opera} proposed OPERA, targeting hallucinations in MLLMs linked to biased self-attention patterns that over-focus on summary tokens, leading to image neglect. OPERA introduces an Over-Trust Logit Penalty in beam search to deprioritize sequences with skewed attention, and a Retrospection-Allocation Strategy to retrospectively adjust token focus and select candidates avoiding hallucinatory patterns, proving effective in reducing hallucinations. From the perspective of visual uncertainty, Leng et al. \cite{leng2023mitigating} introduced VCD, a decoding method addressing object hallucinations in LVLMs by comparing outputs from original and intentionally distorted visuals (the latter is derived by applying pre-defined distortions). This comparison serves as a correct mechanism that helps rectify and refine the model's outputs for accuracy and consistency.

\textbf{Correct Mechanism.} Correction mechanisms typically come in two forms: Post-hoc and In-time. The former involves designing mechanisms (such as leveraging powerful models like GPT-4) to modify the responses generated by LVLMs to achieve more coherent and accurate answers. The latter enhances the context during the model's response generation process to produce more precise answers. 

Post-hoc. Yin et al. \cite{yin2023woodpecker} introduced Woodpecker, a five-part framework using LLMs to correct hallucinations. It extracts key concepts, formulates verification questions, validates with visual knowledge, generates claims at object and attribute levels, and corrects hallucinations using a built knowledge base, enhancing various MLLMs. Similarly, Zhou et al. \cite{zhou2023analyzing} introduced LURE, a method that employs a revisor to correct hallucinatory outputs from LVLMs. They construct a training dataset with GPT-3.5 by (1) adding objects to teach LURE object co-occurrence, and (2) using placeholders for uncertain or concluding objects to prompt re-assessment. This revisor, once trained on this dataset, can be applied to any LVLM to improve response accuracy.

In-time. Zhao et al. \cite{zhao2024mitigating} presented MARINE, a framework that employs DETR \cite{carion2020end} to reduce hallucinations in LVLMs, without the need for further training or APIs. Utilizing a method akin to classifier-free guidance (CFG) \cite{ho2020denoising}, MARINE incorporates DETR tokens during decoding to ensure a balanced consideration of dual tokens, effectively preventing hallucinations.

%% file: Fig_6_hallucination.tex
\begin{figure}[ht]
  \centering
  \includegraphics[width=0.96\textwidth]{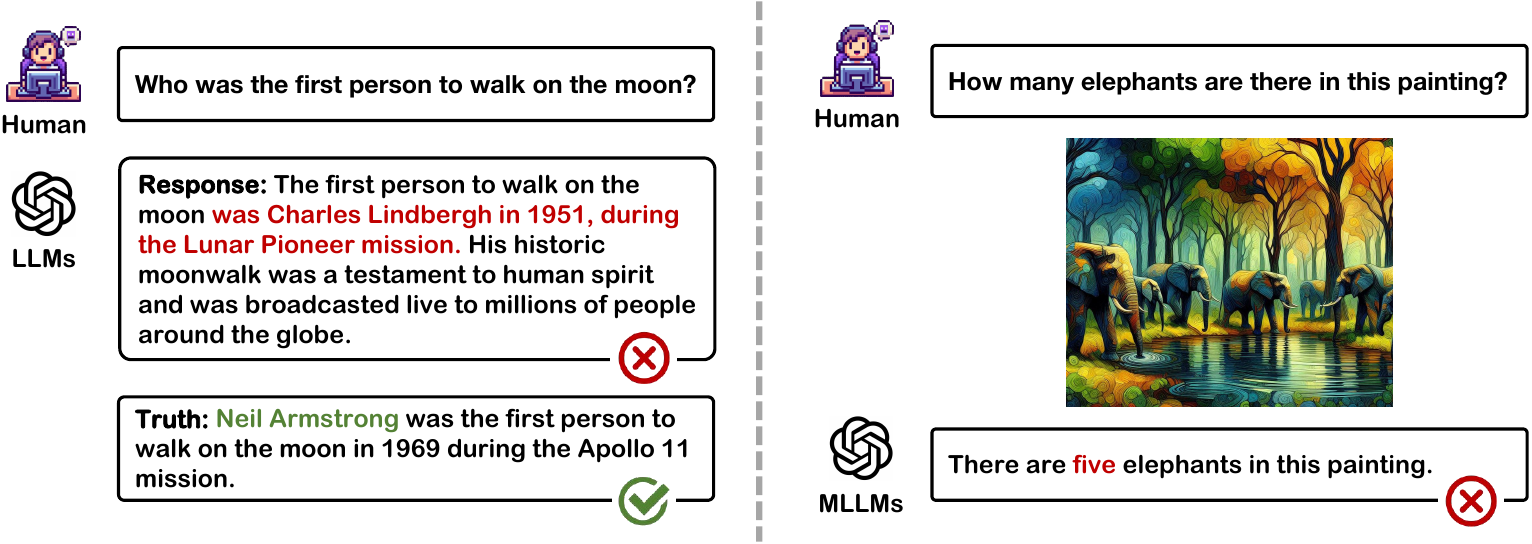} 
  \caption{Examples of LLMs Hallucinations and MLLMs Hallucinations.}
  \label{fig:hallucination} 
\end{figure}

%% file: Fig_9_RAG.tex
\begin{figure}[ht]
  \centering
  \includegraphics[width=0.86\textwidth]{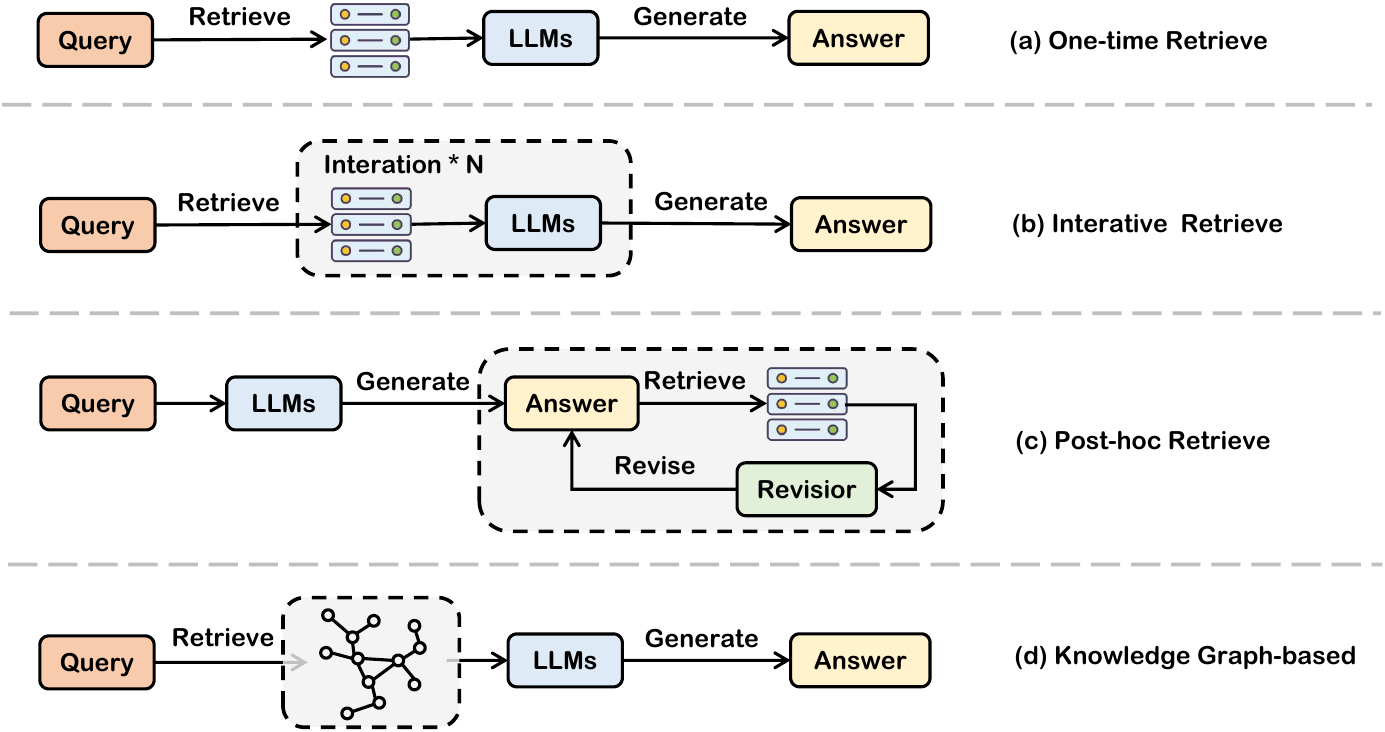} 
  \caption{The illustration of once retrieval, iterative retrieval, post-hoc retrieval, and knowledge graph-based (KG-based).}
  \label{fig: rag} 
\end{figure}

%% file: 5_.tex
\section{FAIGC Detection Methods} \label{sec5:FAIGC Detection methods}
Based on the distinct characteristics of FAIGC content, we categorize existing FAIGC into three types: Deceptive FAIGC, Deepfake, and Hallucination-based FAIGC. In this section, we will introduce the definitions and the prevailing detection methods for FAIGC from the perspectives of the aforementioned three tasks. Refer to Figure \ref{fig:detect tree} for further details.

\input{Fig_10_detect}

\subsection{Deceptive FAIGC Detection} \label{sec5:Deceptive FAIGC detection} 

Deceptive FAIGC is agnostic to the cause of generation. It encompasses AI-generated disinformation created maliciously by users and AI-generated misinformation that arises from hallucinations. Deceptive FAIGC emphasizes the deceitful and harmful nature of the generated content, such as fake news and rumors. This section will discuss the mainstream detection methods for Deceptive FAIGC from both text and multimodal perspectives.

\subsubsection{Text Modality.} \label{sec5:text}

Currently, the detection methods for Deceptive FAIGC in text modality are focused on fake news detection, which often leverages the capabilities of existing LLMs (potentially in conjunction with prompt engineering) or specific models (typically refer to models trained on specific datasets) for detection. Based on traditional neural network models, Zhong et al. \cite{zhong-etal-2020-neural} proposed the use of GNN models, detecting machine-generated fake news by analyzing the text's entity-level consistency and sentence-level consistency. With the introduction of Transformer architecture, early LMs such as GPT \cite{radford2018improving}, BERT \cite{devlin2018bert}, etc., emerged, demonstrating potential in detection tasks. For example, Wang et al. \cite{wang2023implementing} proved through experiments that BERT and Roberta models, trained on a specific dataset, are effective in identifying ChatGPT-generated fake news. Another instance is Zellers et al. \cite{zellers2019defending} trained a model known as Grover, which is capable of generating fake news and indicated through experiments that this generative model itself performs best in detecting fake news. Expanding on this, the emergence of powerful LMs like ChatGPT and GPT-4 provides a fresh perspective for this detection task. Chen et al. \cite{chen2023can} proposed a method leveraging GPT-4 in conjunction with a zero-shot prompt strategy. Experimental results demonstrate that this method outperforms human experts under the LLMFake benchmark, achieving state-of-the-art detection performance and proving that powerful LLMs can serve as efficient and precise detectors. Similarly, Jiang et al. \cite{jiang2023disinformation} suggested the creation of a specific CoT-style prompt to be applied to powerful LLMs to identify machine-generated fake news. Moreover, Wu et al. \cite{wu2023fake} pointed out that when a piece of news undergoes stylistic transformation (rewriting), it can change the judgment of detectors. To address this issue, they proposed a detection framework named SheepDog, which is trained on a specific dataset and emphasizes the consistency of recognition results for different variants of the same piece of news, achieving good detection results.

\subsubsection{Multimodal.} \label{sec5:multimodal}

Similar to the text modality, research on detection methods for Deceptive FAIGC in the multimodal domain also primarily concentrates on fake news detection. For instance, Fung et al. \cite{fung-etal-2021-infosurgeon} proposed InfoSurgeon, a fake news detection framework using knowledge graphs. It initializes node features from text, images, and metadata of multimedia articles, constructs knowledge graphs, and sets KG embeddings. A neural network then learns credibility representations from node connections, facilitating two detection approaches: graph-level classification for overall news authenticity, and binary edge classification to pinpoint specific fake elements. From the perspective of consistency, Tan et al. \cite{tan-etal-2020-detecting} introduced DIDAN, a framework for spotting inconsistencies between images and text in news. It processes Object Representations from images and Caption Word Representations from text, fuses them through an interaction mechanism, and integrates Article Representations for context. The resulting features inform a detector that assigns an authenticity score for identifying fake news.
\subsection{Deepfake Detection} \label{sec5:Deepfake}

Deepfake typically refers to the creation of convincingly realistic fake videos and audio using advanced technologies. In the digital age, the rapid development of Deepfake technology has garnered widespread attention and concern, as the dissemination of such content has sparked social, political, and security issues. Consequently, developing effective Deepfake detection techniques is of paramount importance. In this section, we will provide a comprehensive overview of the current mainstream Deepfake detection methods from both visual and audio perspectives.

\subsubsection{Visual Deepfake} \label{sec5:vision}

Current research in Deepfake detection within the visual modality is centered on the detection of face forgeries \cite{stroebel2023systematic,rana2022deepfake,yu2021survey}. There are primarily two approaches: model-based and feature-based. The former method aims to enhance detection capabilities by improving network architectures, such as integrating multi-attention mechanisms and innovative data augmentation strategies, as well as employing adversarial training to boost generalization ability. The latter approach focuses on utilizing deep features and enhancing the model's adaptability through contrastive learning.

\paragraph{5.2.1.1 Model-based.} \label{sec5:vision model-based} Such methods enhance detection capabilities by improving or proposing new network model architectures. Zhao et al. \cite{zhao2021multi} introduced a deep learning network for detection that features multiple spatial attention heads, texture enhancement blocks, and attention map guidance, along with a new loss function and attention-focused data augmentation. To further refine detection methods, Chen et al. \cite{chen2022self} proposed a method for deep fake detection that improves model robustness by training on a diverse set of fake examples. Using adversarial training to create challenging fakes, their approach demonstrated significant effectiveness in rigorous testing. However, most existing detection algorithms have failed since few visual differences can be observed between an authentic video and a Deepfake one. To address this, Wang et al. \cite{wang2023noise} introduced NoiseDF, a Deepfake detection model that targets forensic noise patterns in Deepfake videos. It combines a RIDNet denoiser with a multi-head interaction method to analyze facial and background noise interactions, showcasing its effectiveness through Deepfake noise trace visualizations. Apart from effectiveness, the generality of detection methods is also critical, as highlighted by Jeong et al. \cite{jeong2022frepgan}, who proposed a deep fake detection framework that employs perturbation maps to render the generated images indistinguishable from authentic ones, thereby enhancing the detector's generality. This model is capable of detecting frequency-level artifacts in early iterations and accounts for image-level irregularities in later stages. Furthermore, leveraging transformer architecture, Coccomini et al. \cite{coccomini2022combining} created a deep fake detection method for facial videos that merges a visual Transformer with EfficientNetB0, matching current top Transformer models. It uses a simple voting mechanism to analyze multiple images without relying on distillation or ensemble methods. 

\paragraph{5.2.1.2 Feature-based.} \label{sec5:vision feature-based} Compared to model-based approaches, this category of methods enhances model capabilities by incorporating mechanisms that leverage internal features to bolster detection efficacy. Current Deepfake detection methodologies typically suffer from limited generalization abilities, struggling to simultaneously detect multiple established manipulation techniques. In response, Cozzolino et al. \cite{cozzolino2023audio} developed POI-Forensics, a multi-modal method leveraging contrastive learning to detect Deepfake through unique audio-visual features of individuals. This approach, focusing on inconsistencies in manipulated embeddings, enhances generalization without requiring fake samples for training. In a parallel effort to enhance the generalizability of detection methods, Dong et al. \cite{dong2023implicit} introduced a Deepfake detection model named ID-unaware to address "Implicit Identity Leakage," where models trained only on binary labels overfocus on identities rather than artifacts, limiting generalization. This method forces the model to pay more attention to local artifacts rather than global identity information, enhancing cross-dataset robustness. Similarly, also focusing on the impact of identity on detection, Huang et al. \cite{huang2023implicit} proposed an implicit identity-driven framework for face swapping detection, which utilizes the distance between explicit and implicit identity features in face images. To extract the required explicit identity features, the framework introduces the explicit identity contrast (EIC) loss and the implicit identity exploration (IIE) loss to supervise the off-the-shelf CNN backbone.

\subsubsection{Audio Deepfake} \label{sec5:Audio}

Currently, the detection of Deepfake in audio modality is primarily focused on determining whether a segment of audio is a product of Deepfake technology \cite{khanjani2023audio,dixit2023review,almutairi2022review}. This can be divided into model-based and metric-based methods. The model-based approach involves enhancing model structures or employing specific models, while the metric-based approach relies on utilizing particular features or designing unique mechanisms. Both strategies aim to improve the detection accuracy of methods as well as to enhance their generalization ability.

\paragraph{5.2.2.1 Model-based.} \label{sec5:model-based} Tak et al. \cite{tak2021end} proposed a detection method for the ANTI-SPOOFING task, which is based on improvements to RawNet2. The enhancements include refraining from applying layer normalization to the input and optimizing the filter length, among others, all intended to adapt RawNet2 for use in anti-spoofing applications. Starting from the standpoint of the underlying causes of Deepfake audio generation, Sun et al. \cite{sun2023ai} developed a detection framework focused on identifying vocoder artifacts, distinctive in synthetic voices but rare in natural audio. The framework trains both a classifier for audio authenticity and a vocoder identification module to refine the feature extractor's focus on vocoder-related artifacts. Furthermore, Kawa et al. \cite{kawa2023improved} proposed a method to harness the feature extraction capabilities of Whisper, integrating the output of the Whisper ASR encoder as a feature with base models such as LCNN and MesoNet for enhanced detection, thereby improving generalization ability.

\paragraph{5.2.2.2 Feature-based.} \label{sec5:feature-based} To tackle the increasingly sophisticated Deepfake production techniques, Krishnan et al. \cite{krishnan2023mfaan} developed MFAAN, a framework leveraging MFCC for timbre, LFCC for spectral characteristics, and Chroma-STFT for harmonics. This combination of audio features facilitates a more nuanced audio analysis, improving the framework's detection capabilities. Similarly, Yang et al. \cite{yang2024robust} proposed a novel multi-feature method. The researchers first demonstrated that learning-based acoustic features have stronger generalization capabilities compared to hand-crafted acoustic features. Building on this, the method introduces two new mechanisms: Feature selection and Feature fusion. The former aims to combine multi-features while avoiding negative features, and the latter aims to integrate multiple features more smoothly. Additionally, Gao et al. \cite{gao2021generalized} suggested the use of long-range spectro-temporal modulation features—2D DCT as artifacts for detection. It can essentially force the CNN classifier to learn from the input audio’s long-term/global modulation patterns, thus improving detection accuracy.

\subsection{Hallucination-based FAIGC Detection} \label{sec5:Hallucination-based FAIGC Detection}

Hallucination-based FAIGC, commonly understood as hallucinations, is widely present in current AI-generated content. They are characterized by their broad coverage across various tasks, diverse forms, and high logical consistency, making them difficult to identify \cite{huang2023survey}. Therefore, there is a need for the design of more efficient and accurate detection methods. In this section, we will elaborate on the current mainstream hallucination detection methods from two perspectives: text and multimodal.

\subsubsection{LLMs Hallucination Detect} \label{sec5:llm}

The detection of hallucinations in text modality can be categorized into two approaches: Grey-box and Black-box methods. Grey-box approaches require full access to the model's output distributions and internal state information (such as values in hidden layers, and self-attention weights). These methods are common in early LMs and are suitable for open-source models. On the other hand, Black-box methods rely solely on the responses of the model. They detect hallucinations by establishing certain metrics or by employing an external discriminator model. Black-box approaches are more widely applicable and can also be used for models that are only accessible through APIs.

\paragraph{5.3.1.1 Grey-box.} \label{sec5:Grey-Box}
Grey-box methods are relatively common in Neural Machine Translation (NMT) tasks. In these tasks, a model's response is based on the source sequence and the target prefix (what has been previously translated at a decoding step). Researchers employ Layer Relevance Propagation (LRP) to analyze the contribution of the source sequence and target prefix to the model's output, which serves as a basis for hallucination detection \cite{voita2020analyzing,ferrando-etal-2022-towards,dale2022detecting}. The central premise of this method is that when a model's output contains hallucination, the contribution of the source sequence to the model's response will be low. Beyond this approach, in NMT tasks, researchers have also detected hallucinations by measuring the uncertainty in the model's responses \cite{fomicheva-etal-2020-unsupervised,zerva-etal-2021-ist}. This method involves assessing the model's confidence through sequence log probability and gauging response uncertainty by comparing the variance in multiple outputs generated via Monte Carlo Dropout. The core idea here is that the higher the uncertainty in the model's response, the higher the likelihood of hallucination. Outside of NMT tasks, for QA tasks, Snyder et al. \cite{snyder2023early} introduced a hallucination detection method using a classifier trained on the internal states of a model, specifically focusing on the initial token's softmax probabilities, Integrated Gradients attributions, self-attention scores, and fully-connected activations to detect hallucinatory responses. Similarly, Azaria et al. \cite{azaria2023internal} introduced a method called SAPLMA, which discerns hallucinations by analyzing the activation values in the model's hidden layers and also involves training a classifier to make this determination. Differing from the previous two methods which measure overall, Luo et al. \cite{luo2023zero} proposed a framework named SELF-FAMILIARITY, which quantifies the model's familiarity with the concepts present in the prompt and uses this as a criterion for detecting hallucinations in the model's answers. The framework prevents the model from generating responses when familiarity falls below a certain threshold.

\paragraph{5.3.1.2 Black-box.} \label{sec5:Black-Box}
Given the current stage where LLMs like GPT-4, Bard, and Claude can only be accessed through APIs and Grey-box methods are not widely and conveniently applicable, Black-box methods become more broadly relevant. Black-box methods can primarily be divided into Token-level Detection and Sentence-level Detection.

\textbf{Token-level Detection.}  Such methods typically discern whether a given token or entity is hallucinated, and are commonly seen in Conditional Sequence Generation tasks. Cao et al. \cite{cao-etal-2022-hallucinated} proposed ENTFA, a hallucination detection method that utilizes the prior and posterior probabilities of an entity based on a pre-trained and fine-tuned masked language model to detect hallucinations. It trains a KNN model to detect the hallucinatory and factual nature of the entity, further categorizing hallucinatory entities into factual and non-factual. For general Conditional Sequence Generation tasks, Zhou et al. \cite{zhou2020detecting} introduced a hallucination detection approach using a classifier trained on a specific dataset. This model assigns a binary label to each position in a model's response, indicating the presence of hallucination. Similarly, Choi et al. \cite{choi-etal-2023-kcts} proposed a classifier-based method for hallucination detection that marks hallucinated tokens by identifying start and inflection points. Tokens between these points are non-hallucinated, while those after the inflection point are considered hallucinated. A trained model is used to determine the start and inflection points. Beyond these works, Dziri et al. \cite{dziri-etal-2021-neural} developed Neural Path Hunter, a tool using knowledge graphs and historical data to identify hallucinations in dialogues generated by LLMs. The detection uses a trained classifier, as a critic, which leverages historical dialogues and knowledge graph data to detect hallucinations in current content.

\textbf{Sentence-level Detection.} These methods typically assess whether an entire sentence contains hallucinations. Some research detects hallucinations by analyzing model uncertainty, as Manakul et al. \cite{manakul2023selfcheckgpt} presented a framework called SelfCheckGPT. This framework evaluates LLM's self-consistency by calculating the similarity among multiple responses generated for the same prompt. Building upon this, Xiong et al. \cite{xiong2023can} proposed an Induced-Consistency evaluation method that assesses model uncertainty by measuring the difficulty with which a model changes its responses when provided with misleading information. Beyond uncertainty, Wang et al. \cite{wang2023hallucination} proposed using search engines to retrieve external knowledge combined with a Bayesian risk decision algorithm for hallucination detection, ensuring the efficiency and accuracy of the detection method. Similarly leveraging external knowledge, Sadat et al. \cite{sadat2023delucionqa} proposed a detection method applied during the RAG process, detecting hallucinations by calculating the similarity and keyword coverage between the LLMs' response and the sentences in the retrieval results. Other research involves using specially trained models or other LMs for hallucination detection. For more detailed detection of hallucinations, Mishra et al. \cite{mishra2024fine} proposed dividing hallucination types into subcategories and training a model named FAVA on a specific dataset to detect these types of hallucinations. Additionally, there are studies focused on specific tasks or scenarios following this idea, such as Zhao et al. \cite{zhao-etal-2023-hallucination} on Grounded Instruction Generation, Shen et al. \cite{shen2023misleading} on news headline detection, Son et al. \cite{son-etal-2022-harim} on identifying hallucinations in summarization tasks, and Mündler et al. \cite{mundler2023self} on detecting self-contradictory hallucinations in models' responses. Beyond targeted detection methods, Zha et al. \cite{zha2023alignscore} introduced a hallucination detection approach applicable to various text-to-text tasks (Natural Language Inference, QA, summarization, etc.).

\subsubsection{MLLM Hallucination Detection} \label{sec5:mllm}

Currently, the primary methods for hallucination detection in MLLMs are targeted at text-to-image and image-to-text (Image Caption) tasks. These methods typically employ prompt engineering, leveraging the capabilities of existing MLLMs or using specific models (refer to specialized models or those trained on designated datasets) for detection purposes. Chen et al. \cite{chen2024unified} presented UNIHD, a four-step hallucination detection framework. Initially, GPT-4V/Gemini extracts claim from image-to-text outputs or decomposes text-to-image queries. Next, it formulates queries from these claims for aspect-oriented tools. These tools then provide insights for verifying hallucinations. Finally, GPT-4V/Gemini assesses each claim using the insights gathered. Inspired by RLHF, Gunjal et al. \cite{gunjal2023detecting} developed a two-tiered reward model to identify LVLM-generated content, focusing on sentence-level integrity and segment-level detail. The model classifies outputs as Accurate or Inaccurate for binary assessment, and adds an Analysis category for ternary classification to capture subjective or non-image-based content. From the perspective of consistency, Yu et al. \cite{yu2023hallucidoctor} developed HalluciDoctor, a framework for detecting and correcting hallucinations in data. For detection, it first identifies answer chunks in responses, prompts LLMs to formulate various questions about these chunks, and gets answers from MLLMs like MiniGPT-4. A Bert-based similarity score (called ConScore) is then calculated between the original and MLLMs' responses. Chunks with a ConScore below 0.5 are flagged as hallucinated. Subsequently, ChatGPT removes these chunks to preserve the context.

%% file: Fig_10_detect.tex
% \definecolor{mycolor}{RGB}{215, 245, 200}

\tikzstyle{my-box}=[
    rectangle,
    draw=black,
    rounded corners,
    text opacity=1,
    minimum height=1.5em,
    minimum width=5em,
    inner sep=2pt,
    align=center,
    fill opacity=.5,
    line width=0.8pt,
]
\tikzset{
leaf/.style={
my-box,
minimum height=1.5em,
fill=mycolor,
text=black,
align=left,
text centered,
inner xsep=2pt,
inner ysep=3pt,
line width=0.8pt
}
}
\begin{figure*}[t!]
    \centering
    \begin{adjustbox}{width=0.95\textwidth}
        \begin{forest}
            forked edges,
            for tree={
                grow=east,
                reversed=true,
                anchor=base west,
                parent anchor=east,
                child anchor=west,
                base=center,
                rectangle,
                draw=black,
                rounded corners,
                align=left,
                text centered,
                minimum width=5em,
                edge+={darkgray, line width=1pt},
                % s sep=3pt,
                inner xsep=2pt,
                inner ysep=3pt,
                line width=0.8pt,
                ver/.style={rotate=90, child anchor=north, parent anchor=south, anchor=center},
            },
            where level=1{text width=12em,font=\normalsize,}{},
            where level=2{text width=11em,font=\normalsize,}{},
            where level=3{text width=8em,font=\normalsize,}{},
            % where level=5{text width=12em,font=\normalsize,}{},
            [FAIGC Detection \\ Methods (Sec. \ref{sec5:FAIGC Detection methods})
               [Deceptive FAIGC \\ Detection(Sec.\ref{sec5:Deceptive FAIGC detection})
                    [Text Modality\\ (Sec.\ref{sec5:text})
                    ] 
                    [Multimodal\\ (Sec.\ref{sec5:multimodal})
                    ]
                ]
                    [Deepfake Detection \\ (Sec.\ref{sec5:Deepfake})
                        [Vision Deepfake \\
                        (Sec.\ref{sec5:vision}),
                        [Model-based\\
                        (Sec.5.2.1.1), leaf
                        ]
                        [Feature-based\\
                        (Sec.5.2.1.2), leaf
                        ]
                        ]
                        [Audio Deepfake\\
                        (Sec.\ref{sec5:Audio}),
                        [Model-based\\
                        (Sec.5.2.2.1), leaf
                        ]
                        [Feature-based\\
                        (Sec.5.2.2.2), leaf
                        ]
                        ]
                    ]
                    [Hallucination-based \\ FAIGC Detection 
                    (Sec.\ref{sec5:Hallucination-based FAIGC Detection})
                        [LLMs Hallucinations \\Detection
                        (Sec.\ref{sec5:llm}),
                        [Grey-Box \\
                        (Sec.5.3.1.1), leaf
                        ]
                        [Black-Box. \\
                        (Sec.5.3.1.2), leaf
                        ]
                        ]
                        [MLLMs Hallucinations \\Detection.
                        (Sec.\ref{sec5:mllm}),
                        ]
                    ]
                ]
            ]
        \end{forest}
    \end{adjustbox}
        % \vspace{-4mm}
    \caption{The structure of Sec.\ref{sec5:FAIGC Detection methods} FIAGC Detection methods. We categorize the FAIGC Detect into three distinct sub-tasks: Deceptive FAIGC Detect, Deepfake Detect, Hallucination-based FAIGC Detection.}
    \label{fig:detect tree}
\end{figure*}
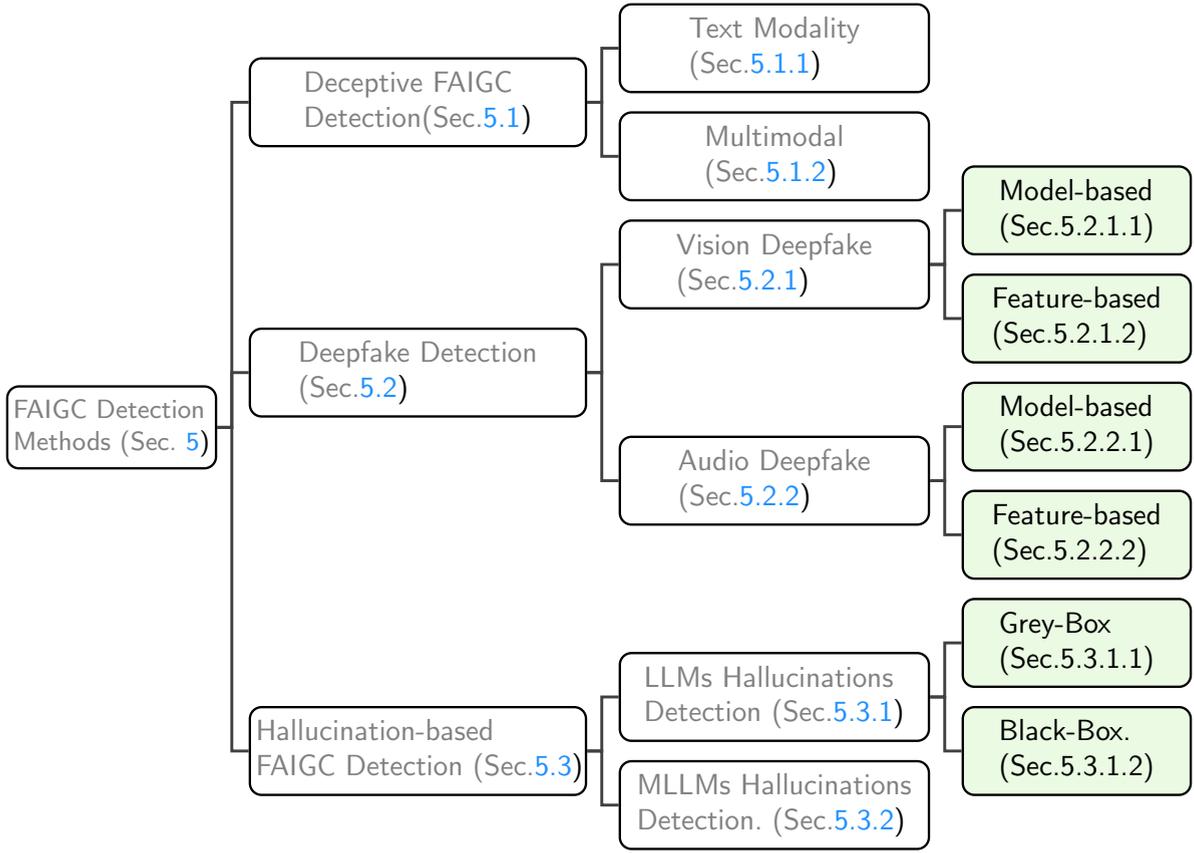

%% file: 6_.tex
\section{FAIGC Detection Benchmark} 

The FAIGC Detection Benchmark is typically designed to evaluate the performance of detection methods. Its corresponding assessment results allow us to understand the performance of detection methods across different tasks, thereby guiding the development and improvement of detection methods, and designing more efficient and comprehensive detection approaches. In this section, we will introduce the prevailing benchmarks for three types of tasks: Deceptive FAIGC, Hallucination-based FAIGC, and Deepfake. All these benchmarks are collected in Table \ref{table: benchmark}.

\input{benchmark}

\subsection{Deceptive FAIGC Detection}

The Deceptive FAIGC detection benchmark focuses on assessing the capacity of detection methods to identify deceptive content within AI-generated responses correctly. It requires detection methods to discern fabricated facts, consistency errors, and similar content. This will test the detection methods' capabilities in factual retrieval, semantic understanding, and more. In this section, we will explore the corresponding benchmarks from both text and multimodal perspectives.

\subsubsection{Text Modality}

The text-related benchmark concentrates on detecting fake news and typically includes an extensive collection of AI-generated news articles. Detection methods are required to discriminate whether these texts are real or to differentiate them from human-written true news.

\textbf{News-style GROVER-generated dataset \cite{zellers2019defending}.} This benchmark consists of two parts: one part is "real news," which includes 10,000 human-written news articles dated before March 2019, and the other part consists of 10,000 news articles generated by GROVER-Mega. Combining these two parts yields a training set of 10,000 samples, with 2,000 for validation and 8,000 for testing. 

\textbf{Webtext-style GPT2-generated dataset \cite{GPT2-generated}.} This benchmark is comprised of two parts: one consists of human-written instances collected from WebText, and the other part consists of Machine-generated instances produced by GPT-2 XL-1542M. This includes a training set of 500,000 samples, a validation set of 10,000, and a test set of 10,000.

\textbf{Fake News dataset \cite{FakeNews}.} This benchmark contains 387,000 various news articles collected from the internet, along with news texts generated by OpenAI's GPT-2 language model.

\textbf{LLMFake \cite{chen2023can}.} This benchmark is themed around fake content generated by LLMs. For the construction process, three disinformation generation methods are utilized: Hallucinated News, Arbitrary, and Controllable Generation. Hallucinated and Arbitrary methods prompt models like ChatGPT to produce fake news. Controllable Generation employs prompts on models, such as Llama and Vicuna, to create disinformation from nonfactual content in Politifact, Gossipcop, and CoAID datasets using techniques like Paraphrasing, Rewriting, Open-ended, and Information Manipulation Generation, with the last one solely using ChatGPT and Politifact. This results in approximately 5,400 samples.

\textbf{UHGEval \cite{liang2023uhgeval}.} This benchmark focuses on the recognition of hallucinated content in the news domain. During the construction process, open-source models are used to generate continuations based on given news headlines and openings, which typically contain hallucinated content. Subsequently, GPT-3.5 is used to extract keywords from the response, GPT-4 is used to determine the correctness of these keywords, and then undergoes a manual review process. Ultimately resulting in approximately 5,000 samples.

\subsubsection{Multimdoal}

The multimodal-related benchmark focuses on detecting fake news. In terms of content and evaluation format, it is similar to the text-related benchmark. However, the distinction lies in that the multimodal benchmark also includes AI-generated fake images or real images, posing higher demands on the detection methods' capabilities.

\textbf{VOA-KG2txt \cite{fung-etal-2021-infosurgeon}.} This benchmark includes a substantial dataset of multimedia fake news, characterized by its fine-grained annotations, which enable the detection of specific elements within the news. The construction process is divided into two methods. The first method involves generating fake news from a KG of a real article, manipulating mid-frequency entities through swapping, adding relations, or replacing subgraphs, and using BART-large for text generation. The second method constructs a KG and uses an Abstract Meaning Representation (AMR) parser for graph manipulation via role switching or predicate negation. This process results in a dataset comprising 15k real news and 15k machine-generated fake news.

\textbf{NeuralNews \cite{tan-etal-2020-detecting}.} This benchmark contains a vast collection of both human and machine-generated articles, along with images and captions. Machine-generated news articles are created by re-writing human articles using models such as GROVER. The data in this benchmark is categorized into four forms: (1) Real Articles with Real Captions, (2) Real Articles with Generated Captions, (3) Generated Articles with Real Captions, and (4) Generated Articles with Generated Captions, each with 32K samples.

\subsection{Hallucination-based FAIGC } 

The hallucination-based FAIGC detection benchmark emphasizes the evaluation of detection methods' ability to correctly identify hallucinated content within AI-generated answers. It requires detection methods to discern factual inaccuracies, directive compliance errors, and similar issues, posing challenges to the detection methods' capabilities in fact-checking and consistency assessment. In this section, we will discuss the corresponding benchmarks from both text and multimodal perspectives.

\subsubsection{LLMs Hallucination}

Text-related benchmarks focus on detecting hallucinations in specific tasks such as QA, RAG, and NLG. Beyond task types, detection is also conducted at the granularity level, including passage-level, sentence-level, and token-level. The combination of these two forms allows for a comprehensive evaluation of the detection methods' performance. Such benchmarks typically include a large amount of AI-generated content within the relevant tasks, and detection methods are required to determine the accuracy of the related facts and whether the instructions have been correctly followed.

\textbf{FactCHD \cite{chen2023unveiling}.} This benchmark focuses on identifying fact-conflicting hallucinated content in LLMs'responses. To construct the benchmark, manually designed queries are used to generate query-response pairs with open-source LLMs, which are annotated as non-factual and factual to serve as examples. Prompts are then designed for ChatGPT to learn from these examples and generate similar query-response pairs, with data selection based on Semantic Diversity. Further prompts lead ChatGPT to generate explanations for these pairs, culminating in a dataset of 6,940 samples after manual curation.

\textbf{PHD \cite{yang-etal-2023-new-benchmark}.} This benchmark focuses on the identification of hallucinated content at the passage level. It is constructed by extracting entities from a Wikipedia dump, dividing them into three categories based on the number of Google Search results (PHD-Low, PHD-Medium, PHD-High), and sampling 100 entities from each category. Prompts guide ChatGPT to generate Wikipedia-style articles, which are manually labeled as factual, non-factual, or unverifiable.

\textbf{AutoHall \cite{cao2023autohall}.} This benchmark is concentrated on the detection of fact-conflicting hallucinations. The construction process involves creating prompts for an LLM to generate references for claims labeled as True or False. The LLM then assesses the claim’s correctness based on the provided claim and reference. If the assessment contradicts the original label, the previously generated reference is considered to contain hallucination. Data for this benchmark are sourced from three known datasets: Climate-fever, Pubhealth, and WICE, resulting in a total of 907, 1009, and 928 samples, respectively.

\textbf{HADES \cite{liu-etal-2022-token}.} This benchmark addresses token-level hallucination detection. It is constructed by sampling text segments from the WIKI-40B dataset, perturbing them with BERT while adhering to fluency, grammatical correctness, and diversity principles. The perturbed texts are manually annotated as hallucination or not, yielding a dataset of 10,954 samples.

\textbf{RAGTruth \cite{wu2023ragtruth}.} This benchmark is focused on hallucination detection during the RAG process. For the construction process, responses are generated for the RAG task using six models, subsequently annotated manually as Yes (hallucination) or No (not hallucination). The RAG tasks are divided into three categories: Summarization, Question Answering, and Data-to-Text generation, with data sourced from CNN/Daily Mail, MS MACRO, and Yelp Open Dataset, respectively, amounting to a total of 17,838 samples.

\textbf{DelucionQA \cite{sadat2023delucionqa}.} This benchmark focuses on hallucination detection in domain-specific QA tasks during the RAG process. It involves using an LLM to generate questions about the Jeep 2023 Gladiator dataset, followed by manual filtering to remove inappropriate questions. A retrieval algorithm provides evidence for these questions, which the LLM uses to generate responses. These responses are manually annotated, resulting in 2,038 samples.

\textbf{HaluEval \cite{HaluEval}.} This benchmark focuses on the identification of hallucinated content within NLG tasks. The construction method is divided into automated generation and human annotation. For the former, using a "sampling-then-filtering" approach prompts ChatGPT to generate hallucinated responses, and then refine them to plausible answers. This generates 10,000 samples each for question answering, text summarization, and knowledge-grounded dialogue. For human annotation, queries are sampled from the 52K instruction tuning dataset from Alpaca, and ChatGPT is used to generate corresponding responses. These are manually labeled as unverifiable, non-factual, and irrelevant, resulting in a dataset of 5,000 samples.

\textbf{SelfCheckGPT-Wikibio \cite{manakul2023selfcheckgpt}.} This benchmark focuses on the detection of fact-conflicting hallucinated content. The construction process involves sampling a set of individuals/concepts from the WikiBio dataset and creating prompts that instruct ChatGPT to generate Wikipedia-style articles. These articles are subsequently manually annotated, with each sentence being labeled as either Major Inaccurate, Minor Inaccurate, or Accurate. This process results in a dataset comprising 1,908 samples.

\textbf{FAVABENCH \cite{mishra2024fine}.} This benchmark concentrates on the detection of hallucinated content within fine-grained factuality tasks. The construction process involves collecting authentic responses from mainstream LLMs across multiple knowledge-intensive tasks. These responses are manually annotated with fine granularity (such as entities, relations, etc.), resulting in a dataset of approximately 1,000 samples.

\subsubsection{Multimodal} 

The multimodal-related benchmark focuses on detecting hallucinations in specific tasks, such as text-to-image and image-to-text (Image Caption) tasks. It typically includes a large volume of AI-generated text or images specific to these tasks and the detection method is required to determine whether the generated content aligns with the visual information or is consistent with the textual requirements.

\textbf{MHaluBench \cite{chen2024unified}.} The benchmark comprises two types of data: text-to-image and image-to-text. In the construction phase, for the former, the dataset includes 200 data points from Image Captioning (IC) tasks and 200 from Visual Question Answering (VQA) tasks. For the latter, it contains 220 data points from Image-to-Text (I2T) tasks. The data is generated by state-of-the-art MLLMs such as LLaVA, MiniGPT-4, and DALLE-3. This benchmark introduces a multi-granular evaluation approach, where the input text and output text are segmented into multiple segments and claims to achieve a more accurate assessment. The annotation is done manually.

\textbf{HalDetect \cite{gunjal2023detecting}.} This benchmark focuses on VQA tasks. The construction process involves sampling 4,000 images from the COCO dataset and generating descriptive responses to those images using the InstructBLIP model. For each image, four responses are sampled using nucleus sampling, resulting in a total of 16,000 samples. The dataset is manually annotated with a fine-grained approach, which entails segmenting each sentence into various fragments and labeling them. The annotations are categorized into four types: Accurate, Inaccurate, Analysis, and Unsure. 'Analysis' represents some form of reasoning or explanatory process, which is typically irrelevant to the provided image.

\subsection{Deepfake Detection}

The Deepfake detection benchmark emphasizes the assessment of detection methods' ability to accurately identify AI-generated Deepfake content. Detection methods are required to discern whether content is Deepfake or to distinguish it from real content. This demands the detection methods to possess robust feature recognition and high generalization capabilities. In this section, we will introduce the corresponding benchmarks from both visual and audio perspectives.

\subsubsection{Visual Deepfake Detect}
The visual-related benchmark focuses on detecting Deepfake content within the specific task of face forgeries. Such benchmarks typically include a large number of Deepfake content generated by face forgery-related techniques in the form of images or videos. Detection methods are required to identify differences between these and authentic content or to detect the altered features within them.

\textbf{CDDB \cite{li2023continual}.} This benchmark extends the stationary Deepfake detection scenario to a dynamic one, in which a stream of likely heterogeneous Deepfake appears over time rather than simultaneously. In this setting, previously encountered Deepfake may become invisible due to certain reasons. This poses a greater challenge to current methods that are more inclined toward static detection. The construction process involves using three types of models: 1) GAN  models, 2) non-GAN models, and 3) unknown models to synthesize images from a series of real sources, ultimately yielding approximately 842k samples.

\textbf{CelebDF \cite{li2020celeb}.} This research introduces CelebDF, a novel and significantly challenging large-scale Deepfake video dataset. Comprising 5,639 high-quality Deepfake videos of celebrities, these videos are crafted using an advanced synthesis technique. The study undertakes an exhaustive evaluation of existing Deepfake detection methodologies and datasets, illustrating the heightened level of challenge that CelebDF presents.

\textbf{FakeAVCeleb \cite{khalid2021fakeavceleb}.} This research introduces FakeAVCeleb, an innovative audiovisual Deepfake dataset. Unlike others, it includes synthetic audio synchronized with the forged visuals, created using prevalent Deepfake methods. The dataset features diverse celebrity videos to enhance realism. The study conducts thorough experiments to evaluate it, revealing challenges and potential detection applications.

\textbf{DF-Platter \cite{narayan2023df}.} This research undertakes a simulation of authentic Deepfake generation scenarios and introduces the DFPlatter dataset. This dataset, with 133,260 videos over 500 GB, includes diverse low and high-res Deepfake of individuals and groups, featuring Indian ethnicities. Its variety and complexity challenge detection models, showing reduced performance, especially with low-res Deepfake. This dataset aims to improve Deepfake detection technology by expanding its scope and effectiveness.

\textbf{VIDEOSHAM \cite{mittal2023video}.} The VIDEOSHAM dataset comprises 826 videos, split evenly between 413 authentic videos and 413 altered videos. While many current Deepfake datasets predominantly focus on two types of facial manipulations—face swapping with different subjects or modifying existing faces—VIDEOSHAM offers a broader range, including rich contextual detail and human-centered high-resolution videos. These videos have been altered using a blend of six distinct spatial and temporal manipulation techniques.

\textbf{Deepfake MNIST+ \cite{huang2021deepfake}.} This research introduces Deepfake MNIST+1, a novel dataset containing 10,000 facial animation videos. It covers ten diverse facial expressions/actions and serves to aid in Deepfake detection. The study includes a baseline detection method and an analysis of its effectiveness. Furthermore, it examines the characteristics of the newly proposed dataset, highlighting the complexities and critical importance of identifying Deepfake videos amidst varying facial expressions and different levels of compression quality.

\textbf{LAV-DF \cite{cai2023glitch}.} This benchmark encompasses a range of strategically designed manipulations in audio, visual, and audio-visual formats driven by content requirements. The purpose of this dataset, LAV-DF, is to address the limitations of current benchmark datasets, which predominantly focus on extensive visual alterations in video content. It uniquely incorporates manipulations that are tactically oriented and content-driven, spanning audio, visual, and combined audio-visual elements.

\subsubsection{Audio Deepfake}

Audio-related benchmark detection tasks primarily assess whether an audio segment belongs to Deepfake content. Such benchmarks typically encompass an extensive collection of Deepfake audio produced by technologies like TTS and VC. Detection methods must be capable of discerning the discrepancies between these and authentic audio content, or determining through specific characteristics whether the audio has been synthesized using Deepfake or analogous technologies.

\textbf{EmoFake \cite{zhao2022emofake}.} This benchmark innovatively proposes the generation of fake audio by modifying the emotional state of the source audio. The construction process involves sampling data from the ESD dataset and utilizing open-source Emotional Voice Conversion (EVC) models, such as VAW-GAN-CWT and DeepEST, to transform the original emotion into another. The result is a dataset comprising 39,900 fake data samples and 17,500 real data samples. 

\textbf{Wavefake \cite{frank2021wavefake}.} This benchmark's construction process starts by sampling data from two datasets, LJSPEECH and JSUT, and uses six state-of-the-art architectures across two languages to generate corresponding recreated fake data. Ultimately, this yields a dataset containing 177,985 generated audio samples.

\textbf{LibriSeVoc \cite{sun2023ai}.} This benchmark focuses on vocoder artifact detection tasks. It is built upon the LibriTTS speech corpus, utilizing six SOTA neural vocoders to generate samples, each synthesized sample will contain specific vocoder artifacts. The resulting dataset includes 13,201 real audio samples and 79,206 fake audio samples.

\textbf{Real-and-fake \cite{reimao2019dataset}.} For this benchmark, the construction process starts with a specific phrase dataset and employs multiple TTS systems to generate synthetic utterances as fake samples. It also collects real human utterances from various open-source datasets and internet recordings to serve as true samples. The final dataset contains more than 87,000 fake samples and more than 111,000 real samples. Depending on the preprocessing method, the dataset can be divided into four subsets: for-rece, for-2-sec, for-norm, and for-original.

\textbf{In-the-Wild \cite{muller2022does}.} This benchmark emphasizes evaluating the generalization capabilities of related work in real-world scenarios. It is constructed by collecting a large amount of audio data from public channels, involving celebrities and politicians. Fake samples are clipped from publicly available Deepfake files, while real samples are gathered by collecting genuine materials such as speeches and podcasts from the same speakers featured in the fake samples. The final dataset includes 19,963 fake samples and 11,816 real samples.

\textbf{ADD2023 \cite{yi2023add}.} This benchmark comprises three subsets: FG-D, LR, and AR. FG-D focuses on the identification of fake utterances, LR on locating the manipulated regions in partially fake audio, and AR aims to recognize the algorithms used in Deepfake utterances. FG-D contains 172,819 real samples and 113,042 fake samples; LR has 46,554 real samples and 65,449 fake samples; AR includes 14,907 real samples and 95,383 fake samples.

\textbf{ASVspoof 2021 \cite{yamagishi2021asvspoof}.} This benchmark is based on the VCTK base corpus and some undisclosed corpora, divided into three subsets: LA, PA, and DF, with the DF subset being relevant to Deepfake detection. The DF subset focuses on the detection of manipulated, compressed speech data posted online, reflecting a scenario where an attacker has access to the voice data of a targeted victim. The DF subset includes 22,617 real samples and 589,212 fake samples.

%% file: benchmark.tex
\begin{table*}[ht]
\centering
\caption{The summary of FAIGC Detection Benchmark.} \label{table: benchmark}
\resizebox{0.98\textwidth}{!}{%
\renewcommand{\arraystretch}{1.5}
\begin{tabular}{c|c|ccccc}
  \toprule[2pt]
  \rowcolor{gray!30}
  
   \textbf{Type} & \textbf{Modality} & \textbf{Benchmark} & \textbf{Datasets} & \textbf{Data Size} & \textbf{Language} & \textbf{Metric} \\
  
   \midrule
  
   \multirow{21}{*}{\textbf{Hallucination-based FAIGC}} &
   \multirow{19}{*}{\textbf{Text}}
   & FactCHD\cite{chen2023unveiling} & - & 6960 & English & F1,ROUGE-L\\
   & & PHD\cite{yang-etal-2023-new-benchmark} & PHD-LOW  & 100 & English & P,R,F1  \\
   & & PHD\cite{yang-etal-2023-new-benchmark}  & PHD-Meidum &100  &  English & P,R,F1 \\ 
   & & PHD\cite{yang-etal-2023-new-benchmark} & PHD-High & 100 & English & P,R,F1 \\ 
   & & AutoHall\cite{cao2023autohall}&  Climate -fever&907  & English & Acc,F1 \\ 
   & & AutoHall\cite{cao2023autohall}& Pubhealth & 1009 & English & Acc,F1 \\ 
   & & AutoHall\cite{cao2023autohall}&  WICE& 928 & English & Acc,F1  \\
   & & HADES\cite{liu-etal-2022-token}& - &  10954& English & P,R,F1 \\
   & & RAGTruth\cite{wu2023ragtruth}& Summarization(CNN/DM) &3768  & English & P,R,F1\\
   & & RAGTruth\cite{wu2023ragtruth}& Summarization(Recent News) & 1896 & English & P,R,F1\\   
   & & RAGTruth\cite{wu2023ragtruth}& Question Answering & 5952 & English & P,R,F1\\
   & & RAGTruth\cite{wu2023ragtruth}& Data-to-text &  6222& English & P,R,F1 \\
   & & DelucionQA\cite{sadat2023delucionqa}& - & 2038 & English & F1 score\\  
   & & HaluEval\cite{HaluEval}&  QA & 10000 & English & Acc \\
   &  & HaluEval\cite{HaluEval}&  Dialogue & 10000 & English & Acc \\
   & & HaluEval\cite{HaluEval}& Summarization   & 10000 & English & Acc  \\
   & & HaluEval\cite{HaluEval}& General   &5000  &  English& Acc \\
   & & FAVABENCH\cite{mishra2024fine}&-  & 1000 & English & F1 score \\
   & & SelfCheckGPT-Wikibio\cite{manakul2023selfcheckgpt}& - & 1908 & English & AUROC   \\

   \cline{2-7}
   
   & \multirow{3}{*}{\textbf{Multimodal}}
   & MHaluBench\cite{chen2024unified}& text-to-image & 400 & English  & P,R,F1,Acc \\
   & & MHaluBench& image-to-text & 220 & English & P,R,F1,Acc \\
   & & M-HalDetect\cite{gunjal2023detecting}& - & 16000 & English & F1,Acc \\

   \midrule

   \multirow{7}{*}{\textbf{Deceptive FAIGC}}
   & \multirow{5}{*}{Text}
   & News-style GROVER-generated dataset\cite{zellers2019defending}& -& 20000 & English & Acc  \\
   & & Webtext-style GPT2-generated dataset\cite{GPT2-generated}&  -& 520000 & English & P,R,F1,Acc \\
   & & Fake News data set\cite{FakeNews}& - &387000  &English  & P,R,F1,Acc \\
   & & UHGEval\cite{liang2023uhgeval}&  -& 5000 & Chinese & Acc,KwPrec\\
   & & LLMFake\cite{chen2023can}& - &5400  &English   & Acc \\

   \cline{2-7}

   & \multirow{2}{*}{\textbf{Multimodal}}
   & VOA-KG2txt\cite{fung-etal-2021-infosurgeon}& - & 30000 & English & F-score,Acc\\
   & & NeuralNews\cite{tan-etal-2020-detecting}& - & 128K & English & Acc \\

   \midrule
   
   \multirow{16}{*}{\textbf{Deepfake}}
   & \multirow{7}{*}{\textbf{Vision}}
   & CDDB\cite{li2023continual}&-  & 842K & - & AA,AF,AA-M,mAP \\
   & & CelebDF\cite{li2020celeb}& - & 6229 &-  & AUC \\
   & & FakeAVCeleb\cite{khalid2021fakeavceleb}& - & 20000 & - & AUC \\
   & & DF-Platter\cite{narayan2023df}& - & 133260 & - & Acc,AUC   \\
   & & VIDEOSHAM\cite{mittal2023video}& - & 826 &-  & DeepFake \\
   & & DeepFake MNIST+\cite{huang2021deepfake}& - &20000 & - & Acc \\
   & & LAV-DF\cite{cai2023glitch}& - &136304  & - & AP,AR \\

   \cline{2-7}

   & \multirow{7}{*}{\textbf{Audio}}
   & EmoFake\cite{zhao2022emofake}& - & 39900 &English  & ERR \\
   & &  Wavefake\cite{frank2021wavefake}& - & 177985 & English & ERR \\
   & & LibriSeVoc\cite{sun2023ai}& - & 92407 & English & ERR \\
   & & Real-and-fake\cite{reimao2019dataset}& - &  195541& English & ERR,Acc  \\
   & & In-the-Wild\cite{muller2022does}& - & 31779 & English & ERR \\
   & & ADD 2023\cite{yi2023add}& FG-D & 285861 & Chinese & F1-score,EER,A-sentence  \\
   & & ADD 2023\cite{yi2023add}& LR & 112003 & Chinese & F1-score,EER,A-sentence \\
   & & ADD 2023\cite{yi2023add}& AR & 110290 &Chinese  & F1-score,EER,A-sentence \\
   & & ASVspoof 2021\cite{yamagishi2021asvspoof}&DF  & 611829 & English & ERR\\

  \bottomrule[2pt]
\end{tabular}
}
\end{table*}

%% file: 7_.tex
\section{Challenges and Future Directions}

There have been notable advancements in the study of FAIGC in recent years, yet there are still some challenges in this field. These challenges not only highlight the current limitations of the research but also indicate possible future research directions. To address these challenges in FAIGC, we have identified several prospective research directions, detailed as follows:

\textbf{Multimodal Deceptive FAIGC Detect.} With the rapid evolution of AIGC technologies, a new challenge emerges: malicious users increasingly exploit AI to craft highly convincing fake multimodal content, thereby significantly complicating public efforts in discerning and counteracting misinformation. Recent research primarily addresses the detection of fake content generated by LLMs in textual form \cite{jiang2023disinformation, wang2023implementing, chern2023factool, goldstein2023generative, pan2023risk}, yet this scrutiny has not been sufficiently extended to multimodal domains. By its complex and realistic nature, Multimodal fake content presents a more daunting detection challenge than its textual counterpart. While some studies have examined the outputs of generative models from a multimodal standpoint \cite{lin2024detecting}, there remains a shortfall in investigating the veracity of these outputs. The multimodal fake content detection field lacks comprehensive research and datasets, which warrants significant academic attention and exploration.

\textbf{Zero-shot FAIGC Detect.} Zero-shot models have shown substantial potential in the fields of AI-generated text detection \cite{mitchell2023detectgpt} and false information detection \cite{lin2023zero}. However, research on detecting FAIGC in a zero-shot learning context remains sparse. Unlike traditional fake information, FAIGC is characterized by a broader variety of production methods and underlying reasons. This diversity poses a significant challenge in creating a comprehensive dataset encompassing different modalities, domains, and FAIGC types, leading to limitations in data resources. We propose a more in-depth investigation into zero-shot learning for the detection of FAIGC, aimed at addressing the issues related to the scarcity of data resources.

\textbf{Interpretable FAIGC Detect.} Several studies have investigated the aspect of interpretability in the field of traditional fake news detection, such as demonstrated in Shu et al. \cite{shu2019defend}. This concept of interpretability can be effectively applied to tasks associated with FAIGC Detect (an acronym that needs clarification in this context). Our perspective is that it's inadequate for a model to determine only if the input data qualifies as FAIGC; it is equally crucial to furnish comprehensive explanations and justifications for these predictions. The requirement for interpretability enhances the model's trustworthiness and, more significantly, furnishes more compelling arguments in the decision-making process.

\textbf{Continual Learning FAIGC Detect Model.} In several instances of hallucinations produced by LLMs, a notable cause is the lack of real-time updates concerning current real-world knowledge \cite{onoe-etal-2022-entity, li-etal-2023-large}. The system referred to as FAIGC exhibits significant real-time capabilities, often incorporating up-to-the-minute information from the real world. To effectively manage the dynamic nature of the real world, models designed for detection purposes must progressively obtain, refresh, compile, and apply knowledge \cite{wu2024continual}. Furthermore, a pressing challenge in research is devising strategies to ensure that these models, while assimilating new knowledge, concurrently retain previously acquired knowledge, thereby preventing the latter's loss.

\textbf{The FAIGC Detection System Constructed through Multi-agent.} Recent studies \cite{talebirad2023multiagent} have unveiled the potential of multi-agent systems in augmenting the capabilities of LLMs. Systems comprising LLMs as agents have demonstrated their proficiency in autonomously handling complex tasks that necessitate a range of human-level skills \cite{chen2023agentverse, liu2023bolaa, xi2023rise}. These systems, through fostering cooperation and exchanging knowledge among agents, markedly enhance the efficiency in tackling complex challenges. Consequently, the application of multi-agent systems in FAIGC detection holds significant promise. Through a consensus mechanism, these systems can amalgamate findings from various detectors to arrive at more substantiated and reliable conclusions.

\textbf{Conducting FAIGC Detection through Fact Verification.} Present-day detection systems exhibit specific limitations in discerning the veracity of FAIGC. Predominantly, these systems emphasize analyzing the characteristic distribution within FAIGC \cite{chern2023factool}. However, they often inadequately address the evaluation of the accuracy of the facts presented in the text. An ideal system for this purpose should proficiently determine the truthfulness of FAIGC content through rigorous fact-checking. Recent research indicates that large language models show considerable potential in identifying factual elements within texts. Furthermore, this capacity for fact verification can be significantly augmented by integrating external knowledge bases \cite{ram2023context} or leveraging search engines \cite{vu2023freshllms,wang2023hallucination}.

%% file: 8_.tex
\section{Conclusion.} 

In the era dominated by AIGC technologies, it is crucial to deeply comprehend the complexities of AI: it significantly boosts societal productivity and brings unprecedented convenience to human life, while also facilitating the emergence of FAIGC. Our survey taxonomizes and discusses FAIGC from three different aspects: the intent behind the FAIGC, the modalities and generative technologies of FAIGC, and the creation methods of FAIGC. It summarizes various technologies involved in FAIGC, detection methods, datasets, and benchmarks. Furthermore, our survey elaborately explores the future research directions of FAIGC. Through this survey, we aim to enhance public and academic awareness of the FAIGC issue and inspire broader and more in-depth research activities to build a healthy, sustainable AI development environment.